\documentclass[letterpaper]{article} 
\usepackage{aaai25}  
\usepackage{times}  
\usepackage{helvet}  
\usepackage{courier}  
\usepackage[hyphens]{url}  
\usepackage{graphicx} 
\usepackage{natbib}  
\usepackage{caption} 
\usepackage{algorithm}

\usepackage{newfloat}
\usepackage{listings}
\DeclareCaptionStyle{ruled}{labelfont=normalfont,labelsep=colon,strut=off} 
\floatstyle{ruled}
\newfloat{listing}{tb}{lst}{}
\floatname{listing}{Listing}
\usepackage{graphicx}
\usepackage{tikz}
\usepackage{amsmath}  
\usepackage{amssymb}  
\usepackage{algpseudocode}  
\usepackage{subcaption}
\usepackage{colortbl}
\usepackage{pifont}
\usepackage{multirow}
\usepackage{textcomp}
\usepackage{adjustbox}
\usepackage{multirow}
\usepackage{booktabs}
\usepackage{hyperref}
\hypersetup{hidelinks,
	colorlinks=true,
        citecolor=black,
        linkcolor=red,
	pdfstartview=Fit,
	breaklinks=true}

\title{EOV-Seg: Efficient Open-Vocabulary Panoptic Segmentation}
\author{
    Hongwei Niu\textsuperscript{\rm 1,\rm 2}\equalcontrib, Jie Hu\textsuperscript{\rm 3}\equalcontrib, Jianghang Lin\textsuperscript{\rm 1}\equalcontrib, Guannan Jiang\textsuperscript{\rm 4}, Shengchuan Zhang\textsuperscript{\rm 1}\thanks{Corresponding Author.}
}
\affiliations{
    \textsuperscript{\rm 1}Key Laboratory of Multimedia Trusted Perception and Efficient Computing,\\
    Ministry of Education of China, Xiamen University, 361005, P.R. China\\
    \textsuperscript{\rm 2}Institute of Artificial Intelligence, Xiamen University, Fujian, China\\
    \textsuperscript{\rm 3}National University of Singapore, Singapore\\
    \textsuperscript{\rm 4}Contemporary Amperex Technology Co., Limited (CATL), Fujian, China\\

    \{niuhw649, hunterjlin007\}@stu.xmu.edu.cn\\
    hujie.cpp@gmail.com, jianggn@catl.com, zsc\_2016@xmu.edu.cn\\
}

\usepackage{bibentry}

\begin{document}
\maketitle
\begin{abstract}
Open-vocabulary panoptic segmentation aims to segment and classify everything in diverse scenes across an unbounded vocabulary.
Existing methods typically employ two-stage or single-stage framework.
The two-stage framework involves cropping the image multiple times using masks generated by a mask generator, followed by feature extraction, while the single-stage framework relies on a heavyweight mask decoder to make up for the lack of spatial position information through self-attention and cross-attention in multiple stacked Transformer blocks.
Both methods incur substantial computational overhead, thereby hindering the efficiency of model inference.
To fill the gap in efficiency, we propose EOV-Seg, a novel single-stage, shared, efficient, and spatial-aware framework designed for open-vocabulary panoptic segmentation.
Specifically, EOV-Seg innovates in two aspects.
First, a Vocabulary-Aware Selection (VAS) module is proposed to improve the semantic comprehension of visual aggregated features and alleviate the feature interaction burden on the mask decoder.
Second, we introduce a Two-way Dynamic Embedding Experts (TDEE), which efficiently utilizes the spatial awareness capabilities of ViT-based CLIP backbone.
To the best of our knowledge, EOV-Seg is the first open-vocabulary panoptic segmentation framework towards efficiency, which runs faster and achieves competitive performance compared with state-of-the-art methods.
Specifically, with COCO training only, EOV-Seg achieves 24.5 PQ, 32.1 mIoU, and 11.6 FPS on the ADE20K dataset and the inference time of EOV-Seg is 4-19 times faster than state-of-the-art methods.
Especially, equipped with ResNet50 backbone, EOV-Seg runs 23.8 FPS with only 71M parameters on a single RTX 3090 GPU.
Code is available at \url{https://github.com/nhw649/EOV-Seg}.
\end{abstract}
\section{Introduction}
\label{sec:intro}
Panoptic segmentation aims to assign a semantic label and a unique instance identifier to each pixel in an image, actually covering both semantic segmentation and instance segmentation.
\begin{figure}[t]
    \begin{minipage}{0.285\linewidth}
    \centerline{\includegraphics[width=\textwidth]{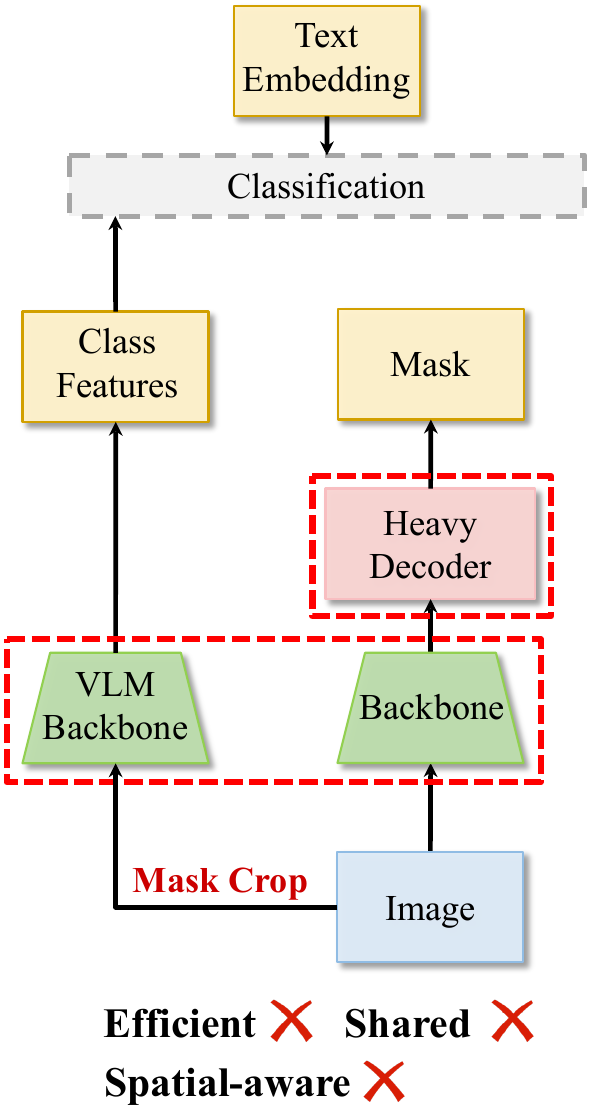}}
        \centerline{(a) Two-stage}
    \end{minipage}
    \begin{tikzpicture}[remember picture, overlay]
        \draw[dashed] (0,2.5) -- (0,-2); 
    \end{tikzpicture}
    \begin{minipage}{0.295\linewidth}
    \centerline{\includegraphics[width=\textwidth]{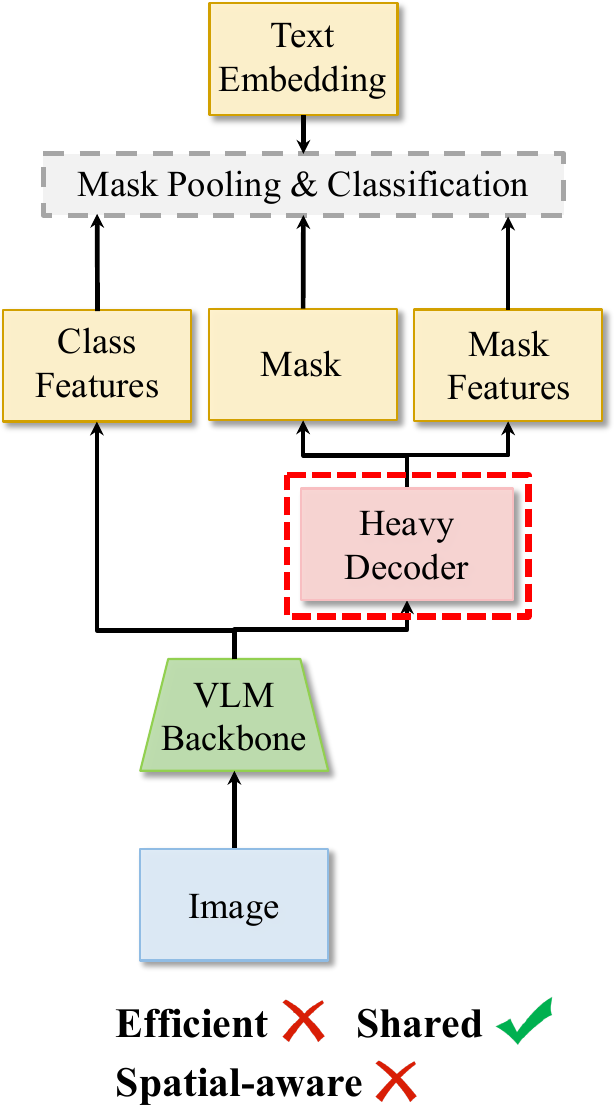}}
        \centerline{(b) Single-stage}
    \end{minipage}
    \begin{tikzpicture}[remember picture, overlay]
        \draw[dashed] (0,2.5) -- (0,-2); 
    \end{tikzpicture}
    \begin{minipage}{0.375\linewidth}
    \centerline{\includegraphics[width=\textwidth]{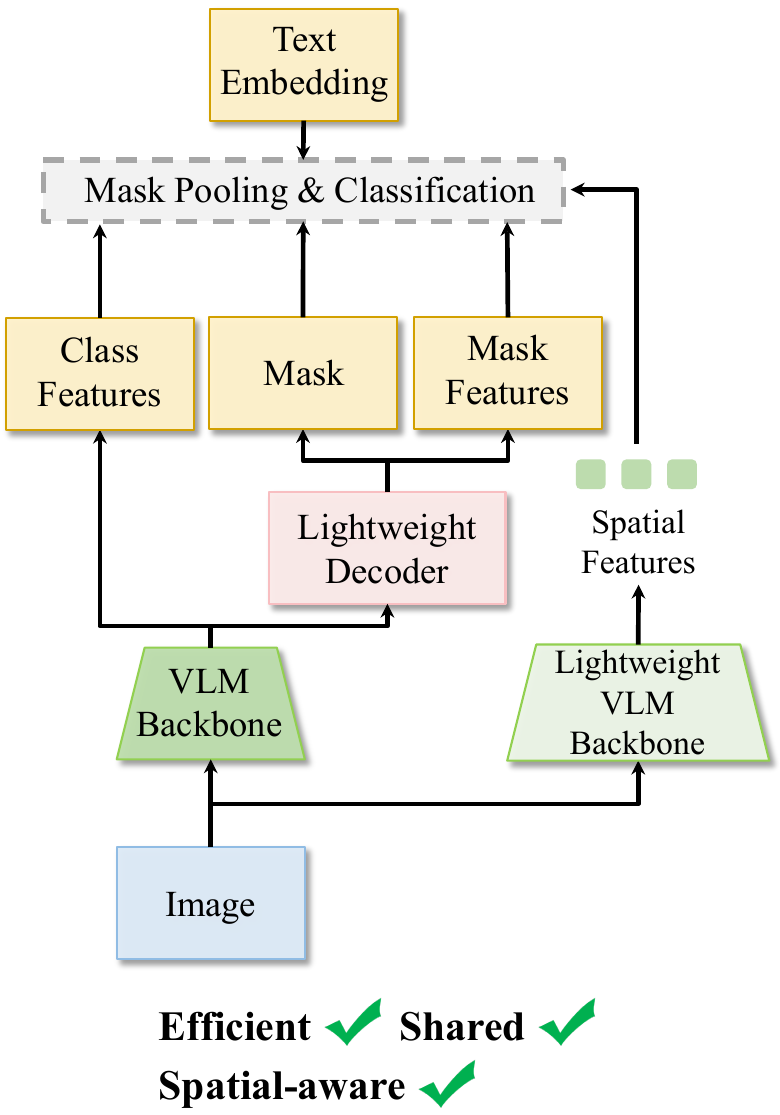}}
        \centerline{(c) EOV-Seg}
    \end{minipage}
    \begin{minipage}{0.47\linewidth}
    \centerline{\includegraphics[width=\textwidth]{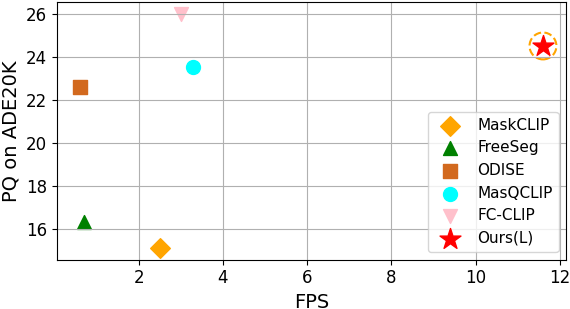}}
        \centerline{(d) Panoptic Segmentation}
    \end{minipage}
    \hspace{0.03\linewidth}
    \begin{minipage}{0.47\linewidth}
    \centerline{\includegraphics[width=\textwidth]{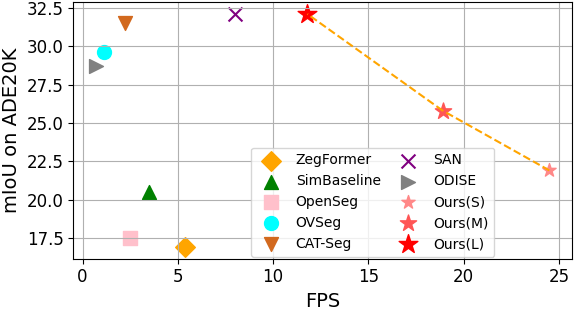}}
        \centerline{(e) Semantic Segmentation}
    \end{minipage}
    \caption{Comparisons between three macro-frameworks for open-vocabulary panoptic segmentation: (a) two-stage, inefficient, non-shared, lack of spatial awareness framework; (b) single-stage, inefficient, shared, lack of spatial awareness framework; (c) Our proposed framework; (d) FPS w.r.t. PQ on the ADE20K datasets for open-vocabulary panoptic segmentation; (e) FPS w.r.t. mIoU on the ADE20K datasets for open-vocabulary semantic segmentation.}
    \label{fig1}
\end{figure}
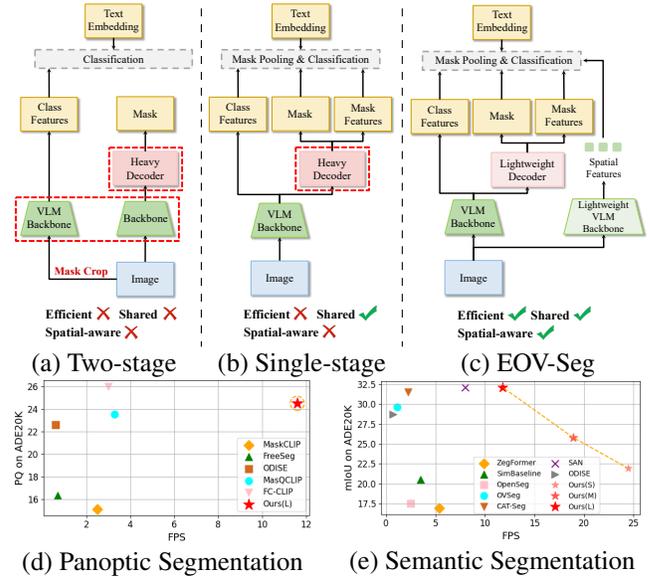
Currently, several methods~\cite{cheng2021mask2former, hu2023you, cavagnero2024pem} have established a unified framework capable of simultaneously addressing semantic, instance, and panoptic segmentation tasks.
However, they are predominantly trained on specific datasets with small-scale predefined categories.
This constraint significantly restricts their adaptability to novel scenarios that feature a diverse and extensive semantic vocabulary.
To overcome this limitation, there has been growing interest in more flexible open-vocabulary segmentation, as explored in works such as~\cite{xu2021simbaseline, liang2023ovseg, qin2023freeseg, ding2023maskclip, 10378471OPSNet, xu2023odise, yu2023fcclip, vs2024possam}.
\begin{figure}[t]
    \begin{subfigure}{0.24\linewidth}
    \centerline{\includegraphics[width=\textwidth]{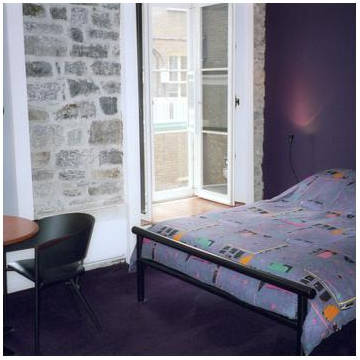}}
        \caption{Input Image}
    \end{subfigure}
    \begin{subfigure}{0.24\linewidth}
    \centerline{\includegraphics[width=\textwidth]{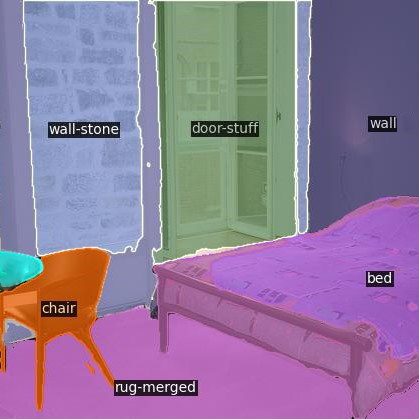}}
        \caption{GT}
    \end{subfigure}
    \begin{subfigure}{0.24\linewidth}
    \centerline{\includegraphics[width=\textwidth]{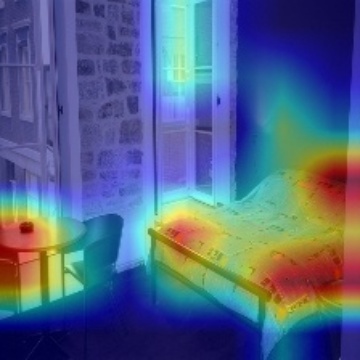}}
        \caption{Grad-CAM}
    \end{subfigure}
    \begin{subfigure}{0.24\linewidth}
    \centerline{\includegraphics[width=\textwidth]{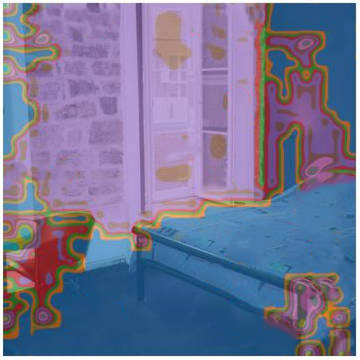}}
        \caption{K-means}
    \end{subfigure}
    \caption{Visualization of Grad-CAM~\cite{jacobgilpytorch-cam} and K-means clustering of CNN-based CLIP~\cite{Radford2021CLIP} backbone features (ConvNeXt-L). (a) Input Image; (b) Ground Truth; (c) Grad-CAM shows the features focusing on local instances; (d) K-means clustering features produce semantically meaningful clusters.}
    \label{fig2}
\end{figure}
These approaches leverage pre-trained Visual-Language Models~(VLMs) like CLIP~\cite{Radford2021CLIP} and ALIGN~\cite{align}.
Despite their innovative nature, these methods still encounter prevalent issues.
Firstly, as shown in Fig.\,\ref{fig1} (a), several methods~\cite{xu2021simbaseline, ding2023maskclip, liang2023ovseg} adopt a two-stage, non-shared, inefficient pipeline that first generates category-agnostic masks, followed by another Vision-Language Model~(VLM) backbone that processes image crops obtained from these masks to extract features for individual classification, which results in high visual feature computational overhead and loss of contextual information.
Secondly, alternatives~\cite{yu2023fcclip, OMGSeg} employ a single-stage, shared, inefficient pipeline, as illustrated in Fig.\,\ref{fig1} (b).
They take a single shared frozen CNN-based CLIP visual encoder as backbone to extract multi-scale features, which is suitable for image segmentation tasks with high-resolution images.
However, as shown in Fig.\,\ref{fig2} (c) and (d), we experimentally observe that the CNN-based CLIP backbone only endows features with the capabilities to discriminate instances while lacking the spatial location capabilities.
This results in that they often employ a heavyweight mask decoder consisting of many transformer layers, including self-attention and cross-attention mechanisms~(self-attention captures contextual information between different queries, whereas cross-attention focuses on specific regions of the feature maps that are relevant to each query to provide awareness of spatial detail) to compensate for the lack of spatial aware capabilities, leading to unacceptable computational overhead and slower inference speed.
In contrast, recent advances in Visual Foundation Models~(VFMs), such as DINOv2~\cite{oquab2023dinov2}, SAM~\cite{kirillov2023sam}, and CLIP~\cite{Radford2021CLIP}, those utilizing Vision Transformer~(ViT)~\cite{dosovitskiy2020vit} architectures as backbones, have demonstrated exceptional zero-shot generalization capabilities.
Through qualitative analysis, as illustrated in Fig.\,\ref{fig3}, we observed nuanced distinctions in the feature representations across different blocks within these models, revealing insights into their discriminative and spatial localization capabilities.
For instance, the clustering of features in the last block of SAM, depicted in Fig.\,\ref{fig3} (c), demonstrates an ability for fine-grained spatial location.
However, the process of sequentially calculating block features results in obtaining these last block features at a significant computational cost.
Conversely, as illustrated in Fig.\,\ref{fig3} (d), the feature clustering of the first block in a ViT-based CLIP model reveals fine-grained object locations with precision comparable to SAM's last block, yet achieves this with significantly less computational overhead.
Therefore, it naturally fits the role of our spatial awareness extractor.
In response to these observations, we introduce EOV-Seg: a novel single-stage, shared, efficient, and spatial-aware framework for open-vocabulary panoptic segmentation as illustrated in Fig.\,\ref{fig1} (c). 
\textbf{1) Single-stage:} We integrate the mask generator and the text embeddings from CLIP text encoder as classifier into a single framework to implement end-to-end open-vocabulary panoptic segmentation.
\textbf{2) Shared:} Instance features extracted by the VLM backbone are shared across classification and mask prediction while keeping image and text features aligned.
\textbf{3) Efficient:} We propose a Vocabulary-Aware Selection~(VAS) module to guide the visual aggregated features to select the features that are more relevant to the texts based on the semantic importance of the texts, thereby improving the semantic comprehension of visual aggregated features and alleviating the feature interaction burden on the mask decoder.
This innovation allows for the deployment of a lightweight decoder as a mask generator, cutting down on computational demands and speeding up the inference process.
\textbf{4) Spatial-aware:} In view of the observed advantages of ViT-based CLIP backbone, we introduce a Two-way Dynamic Embedding Experts~(TDEE) to use weight allocation routers to evaluate the importance of embeddings experts and dynamically allocate expert weights to generate instance embeddings with semantic awareness and spatial awareness for mask recognition.
Extensive experiments show that EOV-Seg runs faster and achieves competitive performance compared with state-of-the-art methods.
In particular, with COCO training only, EOV-Seg achieves 24.5 PQ, 32.1 mIoU, and 11.6 FPS on the ADE20K dataset.
When taking ResNet50 as backbone, it runs 23.8 FPS with only 71M parameters on a single RTX 3090 GPU.
\begin{figure*}[t]
    \begin{minipage}{0.1315\linewidth}
    \centerline{\includegraphics[width=\textwidth]{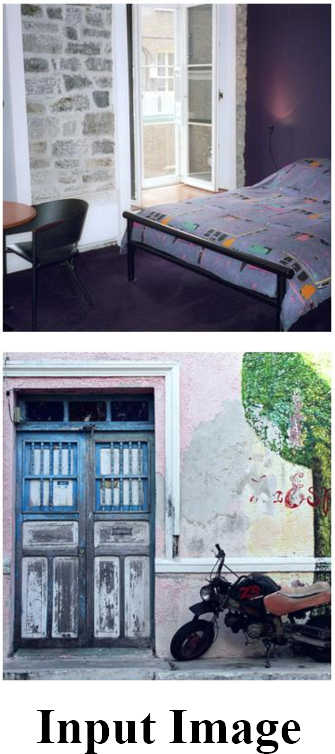}}
        \centerline{(a)}
    \end{minipage}\hfill
    \begin{minipage}{0.27\linewidth}
    \centerline{\includegraphics[width=\textwidth]{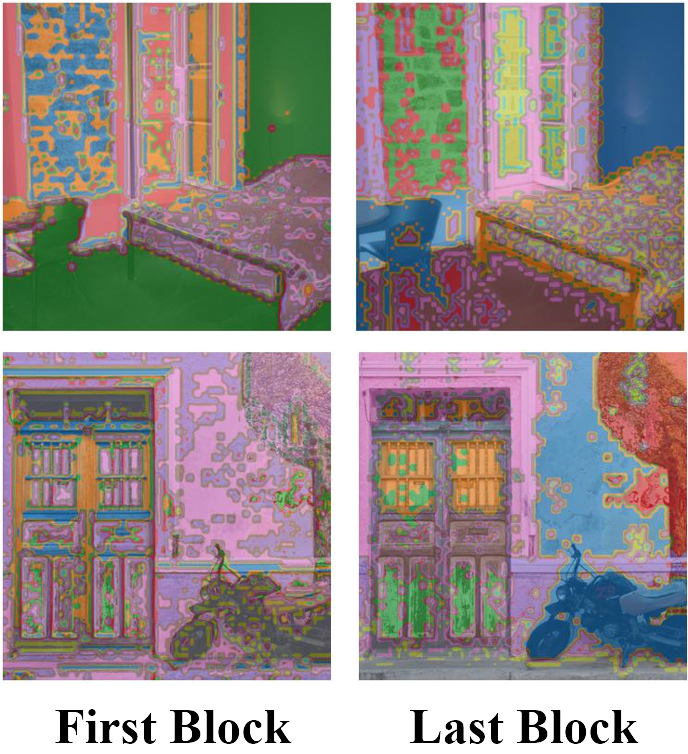}}
        \centerline{(b) DINOv2}
    \end{minipage}\hfill
    \begin{minipage}{0.27\linewidth}
    \centerline{\includegraphics[width=\textwidth]{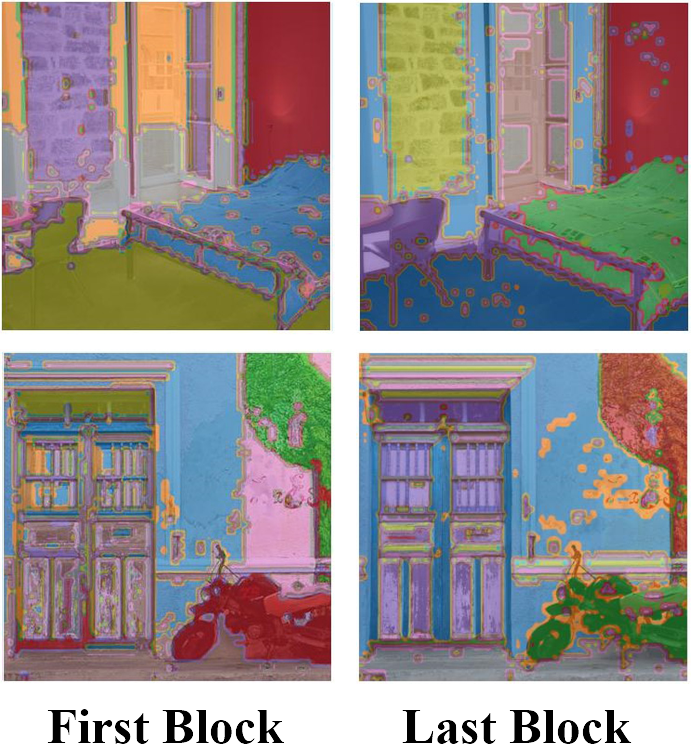}}
        \centerline{(c) SAM}
    \end{minipage}\hfill
    \begin{minipage}{0.27\linewidth}
    \centerline{\includegraphics[width=\textwidth]{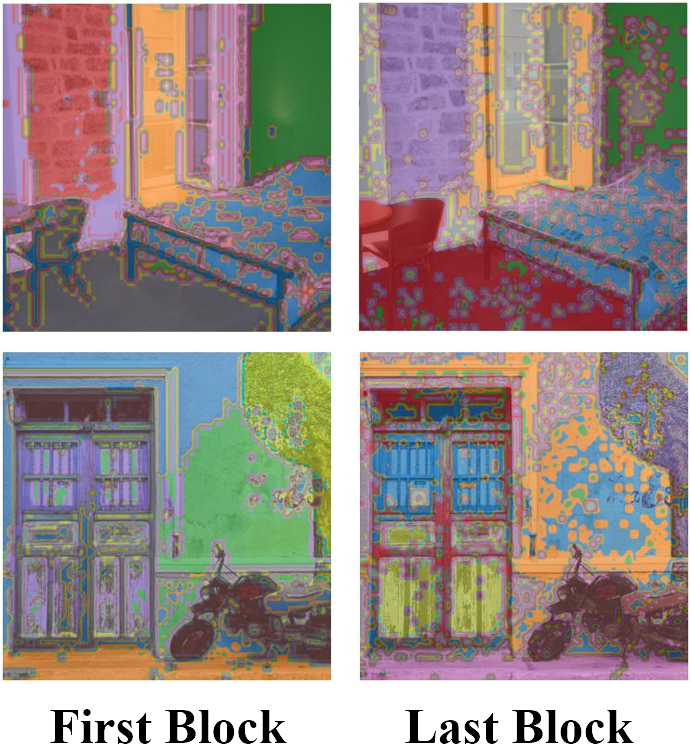}}
        \centerline{(d) ViT-based CLIP}
    \end{minipage}
    \caption{Visualization of K-means clustering of backbone features concerning different blocks across various VFMs.}
    \label{fig3}
\end{figure*}
\section{Related Work}
\label{sec:related}
\textbf{Vision-Language Models.} Vision-language pre-training aims to learn multimodal foundation models with improved performance on various vision-language tasks.
Recently, large-scale Vision-Language Models~(VLMs), such as CLIP~\cite{Radford2021CLIP} and ALIGN~\cite{align}, which are contrastively pretrained on billion-scale, internet-sourced image-text pair datasets, have demonstrated remarkable zero-shot performance on image classification tasks.
With the emergence of large Vision-Language Models~(VLMs), they have been employed across various downstream vision tasks, including object detection~\cite{gu2021vild, zhong2022regionclip, mi2022active, wu2023cora, WSOVOD_2024_AAAI}, image segmentation~\cite{yue2024adaptive}, video understanding~\cite{weng2023Open-VCLIP, hanoona2023vificlip}, 3D scene understanding~\cite{zhang2021pointclip, Zhu2022PointCLIPV2, qu2023sg, qu2024goi, gong2024cross}.
In this paper, we are dedicated to exploring CLIP's powerful instance discrimination and spatial awareness abilities to advance open-vocabulary panoptic segmentation.
\\
\textbf{Open-Vocabulary Panoptic Segmentation.}
%
Open-vocabulary panoptic segmentation is an emerging task in the field of image segmentation, aimed at evaluating the generalization ability to new visual categories that do not exist in the training set. 
Most open-vocabulary panoptic segmentation methods follow a two-stage pipeline.
For example, MaskCLIP~\cite{ding2023maskclip} refines generated masks using transformer-based visual encoder and relative mask attention mechanism, combined with pre-trained CLIP models for panoptic segmentation. 
FreeSeg~\cite{qin2023freeseg} proposes a unified, universal framework to seamlessly handle multiple segmentation tasks through one-time training.
OSPNet~\cite{10378471OPSNet} carefully designs embedding modulation modules and multiple fine-grained components for open-vocabulary segmentation.
In contrast, some methods adopt a single-stage framework.
ODISE~\cite{xu2023odise} uses a text-to-image diffusion model to generate mask proposals and perform classification.
\cite{yu2023fcclip, OMGSeg} builds a single-stage framework by using a frozen CNN-based CLIP backbone, significantly surpassing previous two-stage methods. 
However, these methods suffer from high computational overhead, single-task focus, or lack of spatial location information. 
Our method aims to investigate how to perform efficient open-vocabulary panoptic segmentation with spatial awareness capabilities.
\section{Method}
\label{sec:method}
\textbf{Problem Definition.} Given an input image \(I\in\mathbb{R}^{3\times H\times W}\) and a set of candidate class labels \(C\) \(\in C_{train} \cup C_{test}\).
Open-vocabulary panoptic segmentation aims to segment image~\(I\) into a set of masks~\(M\in \{0, 1\} ^{N\times H\times W}\), each associated with a class label~\(C\), where~\(N\) is the number of masks.
The key challenge of this task arises from training exclusively with classes in~\( C_{train} \), while during inference, \( C_{test} \) may contain classes that are not encountered in~\( C_{train}\).
\begin{figure*}[t]
  \centering
  \includegraphics[width=\textwidth]{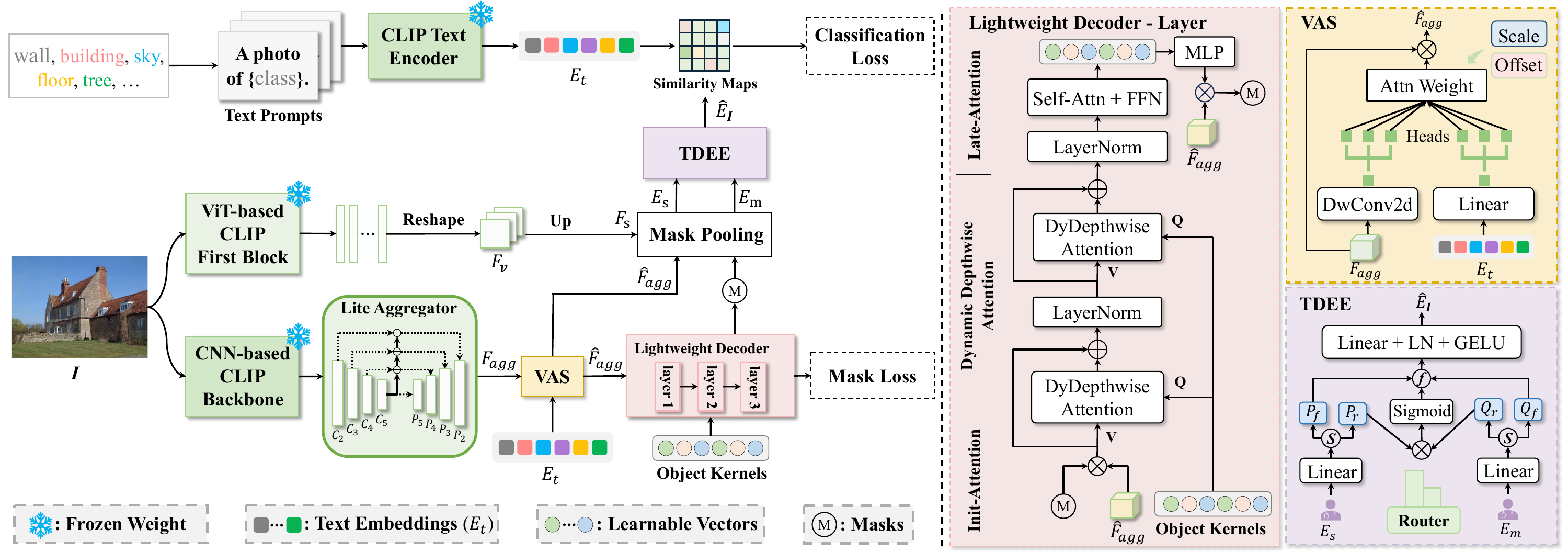}
  \caption{\textbf{Overview of EOV-Seg.} First, the initial block of the ViT-based CLIP is used as a spatial awareness extractor to obtain spatial awareness features \(F_s\). Then, the visual-semantic aggregated features \(\hat{F}_{agg}\) generated by VAS module, the masks generated by the light-weight decoder, and the spatial awareness features \(F_s\) are fed into mask pooling and TDEE sequentially to obtain instance embeddings \(\hat{E}_I\), which will be used to calculate cosine similarity with \(E_t\) for classification.
  }
  \label{fig4}
\end{figure*}
\\
\subsection{Single-Stage, Shared, Efficient Framework}
\label{sec:method_architecture}
Initially, we design a vanilla architecture of the proposed EOV-Seg, which consists of a feature extractor~(CNN-based CLIP backbone), a lite aggregator, a lightweight decoder and a CLIP text encoder.
In the following sections, we detail our elegant designs for elevating the vanilla baseline into the powerful EOV-Seg framework.
\\
\textbf{Feature Extractor.} Since the ViT-based backbone is more sensitive to the input resolution, we take CNN-based CLIP visual encoder as the backbone to extract multi-scale vision features \(C_{i}\), \(i \in \{2, 3, 4, 5\}\) from high-resolution images, with resolutions of \(\frac{1}{4}\), \(\frac{1}{8}\), \(\frac{1}{16}\), and \(\frac{1}{32}\), respectively.
\\
\textbf{Lite Aggregator.} After obtaining multi-scale vision features \(C_{i}\), \(i \in \{2, 3, 4, 5\}\), we use a modulated deformable convolutional~\cite{dai2017deformable_conv} feature pyramid structure to enhance and fuse features from different scales to obtain the intermediate FPN features \(P_{i}\), \(i \in \{2, 3, 4, 5\}\).
To further reduce computational overhead and capture contextual information of different granularities, different scales features are aggregated to obtain the visual aggregated features \(F_{agg}\in\mathbb{R}^{D\times \frac{H}{4}\times \frac{W}{4}}\) as follows:
\begin{align}
F_{agg}=Fuse\left(\sum_{i=3}^N Up_{i}\left(Conv\left(P_{i}\right)\right)+Conv\left(P_{2}\right)\right),
\end{align}
where \(Up_{i}\) indicates an \(2^{(i-2)} \times\) bilinear interpolation operation, \(N\) is number of feature layers, \(Conv\) and \(Fuse\) is $1 \times 1$ and $3 \times 3$ convolutions, respectively, used for adjusting feature dimensions and further fusion of the visual aggregated features \(F_{agg}\).
\\
\textbf{Lightweight Decoder.} As shown in the red part of Fig.\,\ref{fig4}, unlike the heavyweight mask decoder of the Mask2Former~\cite{cheng2021mask2former}, our decoder only consists of three modules at each layer in the following order: a initial-attention module, a dynamic depthwise attention module, and a late-attention module.
Specifically, to further reduce the computational overhead of subsequent feature interactions, the initial-attention module is designed to extract the initial-attention features \(F_{init}\in\mathbb{R}^{N\times D}\) by performing the dot product of the visual-semantic aggregated features \(\hat{F}_{agg}\in\mathbb{R}^{D\times \frac{H}{4}\times \frac{W}{4}}\) from VAS module~(which will be introduced in the following section) with the init masks or masks \(M\in\mathbb{R}^{N\times \frac{H}{4}\times \frac{W}{4}}\) from previous layer.
Next, dynamic depthwise attention is used to conduct cross-dimensional feature interaction between a set of learnable object kernels \(K\in\mathbb{R}^{N\times D}\) and the initial-attention features~\(F_{init}\), producing the refined object kernels \(\hat{K}\in\mathbb{R}^{N\times D}\) as follows:
\begin{align}
\hat{K} &= DyDepthwiseAttention(K, F_{init})
\\
&=F_{init} \ast r(K W_m) \notag
\end{align}
where \(W_m\in\mathbb{R}^{D\times m}\) is a linear layer that projects \(K\) to generate kernels for depthwise convolution, \(r(\cdot)\) is view operation that reshapes the input to $N\times 1\times m$, \(\ast\) denotes the Conv1d operation that takes \(r(\cdot)\) as parameters and $F_{init}$ as input.
The refined object kernels \(\hat{K}\) further concentrate on the relation between different object kernels to enrich information by a multi-head self-attention and a feed-forward network~(FFN).
These refined object kernels \(\hat{K}\) function similarly to a Region Proposal Network~\cite{renNIPS15fasterrcnn}, offering the potential to generate mask proposals.
%
%
Then, we mapped the refined object kernels \(\hat{K}\) to be mask kernels through a three-layer MLP.
%
Each binary mask \(M_i\in\mathbb{R}^{H\times W}\) can be obtained by performing a dot product between the \(i^{th}\) mask kernel and the visual-semantic aggregated features \(\hat{F}_{agg}\).
%
Additionally, the mask embeddings \(E_{m}\in\mathbb{R}^{N\times D}\) is obtained by mask pooling over the visual-semantic aggregated features \(\hat{F}_{agg}\) with predicted masks \(M\).
\\
\textbf{Text Encoder.} Given N class labels, we follow the precedent of prompt engineering~\cite{gu2021vild} by providing M prompt templates for each class, such as ``a photo of a \{class\}.". 
We use a frozen CLIP text encoder to encode all text prompts and then average them across the dimension of the number of templates to obtain the text embeddings \(E_t \in \mathbb{R}^{N_{class} \times D}\).
\subsection{Vocabulary-Aware Selection}
\label{sec:method_VAS}
To alleviate the feature interaction burden on the decoder and facilitate the deployment of a light-weight decoder, we propose Vocabulary-Aware Selection~(VAS) as illustrated in the yellow part of Fig.\,\ref{fig4}.
It can guide the selection of image features based on the semantic importance of the text, thereby aggregating features that are more relevant to the input text and enhancing the semantic understanding of visual features.
Specifically, the visual aggregated features \(F_{agg}\) from the Lite Aggregator and the text embeddings \(E_t\) from the text encoder are mapped to the same feature dimension through linear layers, respective.
Next, they are divided into multi-head visual aggregated features \(\ddot{F}_{agg}\in\mathbb{R}^{h\times \frac{D}{h}\times  \frac{H}{4}\times \frac{W}{4}}\) and multi-head text embeddings \(\ddot{E_t}\in \mathbb{R}^{N_{class} \times h\times \frac{D}{h}}\) along the channel dimension, where \(h\) denotes the number of heads.
This allows us to process complex semantic feature relationships in different subspaces.
Then, they are processed through matrix operations to obtain multi-head attention \(A\in\mathbb{R}^{h\times \frac{H}{4}\times \frac{W}{4}\times N_{class}}\), which is then subjected to a softmax function to produce smoothed logits.
Following this, the maximum value is selected across the vocabulary dimension to acquire vocabulary-aware attention weights \(\ddot{A}\in\mathbb{R}^{h\times \frac{H}{4}\times \frac{W}{4}}\):
\begin{equation}
    \ddot{A} = Max^{(-1)}(Softmax^{(-1)}(\sum^{D/h}_{d} \ddot{F}_{agg} \cdot \ddot{E_t})),
\end{equation}
where \((-1)\) denotes the vocabulary dimension.
Additionally, two learnable parameters, namely scale factor \(\gamma\) and offset factor \(\delta\), are introduced to enable the network to adaptively adjust the vocabulary-aware attention weights \(\ddot{A}\) under various semantic distributions.
Finally, the vocabulary-aware attention weights \(\ddot{A}\) are extended along the channel dimension, followed by element-wise multiplication with the multi-head aggregated features \(\ddot{F}_{agg}\).
The weighted features are then reshaped into the original aggregated features dimension to obtain the visual-semantic aggregated features \(\hat{F}_{agg}\in\mathbb{R}^{D\times \frac{H}{4}\times \frac{W}{4}}\):
\begin{equation}
    \hat{F}_{agg} = Reshape((\gamma \cdot \ddot{A} + \delta) \cdot \ddot{F}_{agg}),
\end{equation}
\subsection{Two-way Dynamic Embedding Experts}
\label{sec:method_TDEE}
According to the analysis in the introduction section, spatial location information is essential for features that possess only instance-level discrimination.
To efficiently extract spatial location information, we take the first block of the CLIP ViT-B visual encoder as spatial awareness extractor.
Specially, the input image \(I\) is processed by the first block of ViT-B \(V_{b_{0}}\) to generate visual tokens (excluding the class token), which are then reshaped to two-dimensional spatial dimensions to obtain the visual-spatial features \(F_v\).
Subsequently, we employ two transposed convolutions for \(4\times\) upsampling to obtain the spatial awareness features \(F_{s}\in\mathbb{R}^{D\times \frac{H}{4} \times \frac{W}{4}}\).
Similar to mask embeddings \(E_m\), mask pooling is used to extract regions of interest from spatial awareness features \(F_s\) as the spatial awareness embeddings \(E_{s}\in\mathbb{R}^{N\times D}\).
Moreover, we propose Two-way Dynamic Embedding Experts~(TDEE) as illustrated in the purple part of Fig.\,\ref{fig4}, which can adaptively incorporate spatial location information and instance-level discriminative information into mask embeddings to improve mask recognition capabilities under different data distributions and a wide range of semantic categories.
Inspired by~\cite{6797059moe}, we model the mask embeddings \(E_m\) and the spatial awareness embeddings \(E_s\) as experts with spatial awareness and instance-level discrimination capabilities, respectively.
They should fulfill their respective roles while also synergistically collaborating.
However, directly correlating the two using fixed weight coefficients prevents the network from perceiving the varying importance of embeddings, leading to either treating all embeddings equally or introducing bias.
Therefore, a weight allocation router is introduced to estimate the significance of the current expert embeddings and adaptively adjust the correlation weight coefficients, thereby facilitating a better integration between the two.
Specifically, the dynamic parameters \(P\in\mathbb{R}^{N\times d}\) for the mask embeddings \(E_m\) and \(Q\in\mathbb{R}^{N\times d}\) for the spatial awareness embeddings \(E_s\) are generated through linear layers, respectively.
The dynamic parameters are evenly split along the channel dimension into router parameters \(P_{r}\in\mathbb{R}^{N\times d/2}\) and \(Q_{r}\in\mathbb{R}^{N\times d/2}\), and fusion parameters \(P_{f}\in\mathbb{R}^{N\times d/2}\) and \(Q_{f}\in\mathbb{R}^{N\times d/2}\).
Next, total router parameters \(P_{t}\in\mathbb{R}^{N\times d/2}\) are aggregated via an element-wise product between \(P_{r}\) and \(Q_{r}\).
Two router linear layers followed by sigmoid activation functions are then employed to provide the experts with adaptive weighting ability, as follows:
\begin{align}
(P_{f}&, P_{r}) = s(E_m W_m), \\
(Q_{f}&, Q_{r}) = s(E_s W_s), \\
P_{t} &= P_{r} \cdot Q_{r}, \\
\alpha_m &= \sigma(LN(Linear_{m}(P_t))), \\
\alpha_s &= \sigma(LN(Linear_{s}(P_t))),
\end{align}
where \(W_{m}\in\mathbb{R}^{D\times d}\), \(W_{s}\in\mathbb{R}^{D\times d}\) project \(E_m\) and \(E_s\) to generate dynamic parameters, respectively.
\(s(\cdot)\) is split operation along the channel dimension.
\(\sigma\) is sigmoid activation function, \(LN\) indicates layer normalization.
This enables the router to adaptively assign weights to the embeddings experts.
%
Finally, a weighted summation is performed as follows:
\begin{align}
\hat{E}_I = \alpha_m \cdot LN(P_{f}) + \alpha_s \cdot LN(Q_{f}),
\end{align}
To further aggregate, a linear layer, LayerNorm (LN), and GELU are employed to obtain the instance embeddings \(\hat{E}_I\).
In general, TDEE takes advantage of the weight allocation router to adaptively assign individual weights to each embeddings expert based on the significance of the embeddings.
As a result, the spatial awareness embeddings \(E_s\), which have strong spatial awareness capabilities, are integrated with the mask embeddings \(E_m\), which contain only instance discrimination information. 
This integration improves the spatial and semantic understanding of the model in open scenes.
\begin{table*}[t]
    \centering
    \begin{adjustbox}{max width=\textwidth}
    \begin{tabular}{l|c|c|c|c|c|c|c|c|c|c|c}
    \toprule
    Method & VLM & Backbone & Training datasets & PQ & SQ & RQ & AP & mIoU & FPS $\uparrow$ & Params$\downarrow$ & FLOPs$\downarrow$\\
    \midrule
    OPSNet~\cite{10378471OPSNet} & ResNet50 & Swin-L & COCO Panoptic+IM-21K & 19.0 & 52.4 & 23.0 & - & - & - & - & 485G\\
    ODISE~\cite{xu2023odise} & ViT-L/14 & Stable Diffusion & COCO Panoptic & 22.6 & 65.1 & 27.0 & 14.4 & 29.9 & 0.6 & 1522M & 953G\\
    ODISE~\cite{xu2023odise} & ViT-L/14 & Stable Diffusion & COCO Panoptic+Caption & 23.4 & 78.1 & 28.3 & 13.9 & 28.7 & 0.6 & 1522M & 953G\\
    FreeSeg~\cite{qin2023freeseg}& ViT-B/16 & ResNet101 & COCO-Stuff-156 & 16.3 & 71.8 & 21.6 & 6.5 & 24.6 & 0.7 & 219M & -\\
    MaskCLIP~\cite{ding2023maskclip} & ViT-L/14@336px & ResNet50 & COCO Panoptic & 15.1 & 70.5 & 19.2 & 6.0 & 23.7 & 2.5 & 367M & 542G\\
    FC-CLIP\textsuperscript{\textdagger}~\cite{yu2023fcclip} & ConvNeXt-L & shared & COCO Panoptic & 26.0 & 70.6 & 31.3 & 16.4 & 33.6 & 3.0 & 221M & 381G\\
    MasQCLIP~\cite{xu2023masqclip} & ViT-L/14@336px & ResNet50 & COCO Panoptic & 23.5 & 69.6 & 29.2 & 12.8 & 30.3 & 3.3 & 373M & 357G\\
    \midrule
    \midrule
    \rowcolor{gray!10} EOV-Seg (L) & ConvNeXt-L & shared & COCO Panoptic & 24.5 & 70.2 & 30.1 & 13.7 & 32.1 & 11.6 & 225M & 377G\\
    \rowcolor{gray!10} EOV-Seg (M) & ResNet50x4 & shared & COCO Panoptic & 18.7 & 63.5 & 23.2 & 8.5 & 25.5 & 18.4 & 127M & 176G\\
    \rowcolor{gray!10} EOV-Seg (S) & ResNet50 & shared & COCO Panoptic & 15.1 & 57.0 & 18.9 & 7.2 & 21.9 & 23.8 & 71M & 133G\\
    \bottomrule
    \end{tabular}
    \end{adjustbox}
    \caption{Open-vocabulary panoptic segmentation performance for training on COCO and testing on ADE20K. ``\textdagger'' means the results are from a re-implementation of FC-CLIP~\cite{yu2023fcclip}. ``$\uparrow$'' and ``$\downarrow$'' means the larger or smaller the value, the better. ``IM-21K'' denotes ImageNet-21K datasets.}
    \label{tab1:adk20_panoptic}
\end{table*}
\begin{table*}[t]
    \begin{adjustbox}{max width=\textwidth}
    \begin{tabular}{l|c|c|c|c|c|c|c|c|c|c|c}
    \toprule
    Method & VLM & Backbone & Training datasets & A-847 & PC-459 & A-150 & PC-59 & PAS-20 & FPS $\uparrow$ & Params$\downarrow$ & FLOPs$\downarrow$\\
    \midrule
    GroupViT~\cite{xu2022groupvit}& ViT-S/16 & ViT-S/16 & GCC+YFCC & 4.3 & 4.9 & 10.6 & 25.9 & 50.7 & - & - & -\\
    LSeg+~\cite{ghiasi2021OpenSeg_LSeg+}& ALIGN & ResNet101 & COCO-Stuff & 2.5 & 5.2 & 13.0 & 36.0 & 59.0 & - & - & -\\
    OpenSeg~\cite{ghiasi2021OpenSeg_LSeg+} & ALIGN & ResNet101 & COCO Panoptic+LN & 4.4 & 7.9 & 17.5 & 40.1 & - & - & - & -\\
    OVSeg~\cite{liang2023ovseg} & ViT-L/14 & Swin-B & COCO-Stuff+Caption & 9.0 & 12.4 & 29.6 & 55.7 & 94.5 & 1.1 & 531M & 830G\\
    CAT-Seg~\cite{cho2024cat-seg} & ViT-L/14 & Swin-B & COCO-Stuff & 10.8 & 20.4 & 31.5 & 62.0 & 96.6 & 2.2 & 490M & 473G\\
    DeOP~\cite{Han2023DeOP} & ViT-B/16 & ResNet101c & COCO-Stuff-156 & 7.1 & 9.4 & 22.9 & 48.8 & 91.7 & 2.5 & 211M & 480G\\
    OVSeg~\cite{liang2023ovseg} & ViT-B/16 & ResNet101c & COCO-Stuff+Caption & 7.1 & 11.0 & 24.8 & 53.3 & 92.6 & 3.4 & 211M & 830G\\
    ZegFormer~\cite{ding2021ZegFormer} & ViT-B/16 & ResNet101 & COCO-Stuff-156 & 4.9 & 9.1 & 16.9 & 42.8 & 86.2 & 5.4 & 211M & 456G\\
    ZegFormer~\cite{ding2021ZegFormer} & ViT-B/16 & ResNet101 & COCO-Stuff & 5.6 & 10.4 & 18.0 & 45.5 & 89.5 & 5.4 & 211M & 456G\\
    SAN~\cite{xu2023SAN} & ViT-L/14 & ViT-liked & COCO-Stuff & 12.4 & 15.7 & 32.1 & 57.7 & 94.6 & 8.0 & 437M & 403G\\
    \midrule
    \rowcolor{gray!10} EOV-Seg (L) & ConvNeXt-L & shared & COCO Panoptic & 12.8 & 16.8 & 32.1 & 56.9 & 94.8 & 11.8 & 225M & 377G\\
    \rowcolor{gray!10} EOV-Seg (M) & ResNet50x4 & shared & COCO Panoptic & 7.8 & 12.2 & 25.5 & 51.8 & 91.2 & 18.9 & 127M & 176G\\
    \rowcolor{gray!10} EOV-Seg (S) & ResNet50 & shared & COCO Panoptic & 6.6 & 11.5 & 21.9 & 46.0 & 87.2 & 24.5 & 71M & 133G\\
    \bottomrule
    \end{tabular}
    \end{adjustbox}
    \caption{Open-vocabulary semantic segmentation performance. We report the mIoU results on five widely used test sets for open-vocabulary semantic segmentation. ``$\uparrow$'' and ``$\downarrow$'' means the larger or smaller the value, the better. ``LN'' denotes Localized Narrative datasets.}
    \label{tab2:adk20_semantic}
\end{table*}
\section{Experiments}
\label{sec:exp}
\subsection{Implementation Details}
\label{subsec:exp_impl}
\textbf{Architecture.}
We use a CNN-based CLIP~\cite{Radford2021CLIP} visual encoder as backbone, S, M, L denote ResNet50, ResNet50x4, ConvNeXt-Large, respectively.
Additionally, we utilize the first block of the ViT-B/16 CLIP model as a spatial awareness extractor. 
\\
\textbf{Training Strategy.}
Following prior works ~\cite{xu2023odise,yu2023fcclip}, the training batch size is 16.
We train our EOV-Seg on 4 NVIDIA 3090 GPUs for a total of 200k iterations.
We only use the COCO Panoptic~\cite{lin2014coco} dataset for training, with a crop size of 1024$\times$1024.
All of our experiments are performed three times and averaged for fair comparison.
\\
\textbf{Evaluation Protocols.}
We evaluated our EOV-Seg for open-vocabulary semantic, instance, and panoptic segmentation on the ADE20K~\cite{zhou2017ade20k} dataset, and for semantic segmentation on the ADE20K~\cite{zhou2017ade20k}, PASCAL Context~\cite{6909514pascalvoc2} and PASCAL VOC~\cite{Everingham10pascalvoc1} datasets.
During inference, the shortest side of input images is resized to 640, while ensuring the longer side does not exceed 2560. 
\textit{See Appendix for more details.}
\subsection{Main Results}
\textbf{Open-Vocabulary Panoptic Segmentation.} 
Tab.\,\ref{tab1:adk20_panoptic} reports the performance of different open vocabulary panoptic segmentation methods on the ADE20k dataset.
Compared to existing state-of-the-art methods, EOV-Seg (L) runs faster and achieves competitive performance.
Specifically, although ODISE~\cite{xu2023odise} take powerful text-to-image diffusion~\cite{stable_diffusion} model as backbone and is trained on the additional COCO Caption~\cite{chen2015coco_caption} dataset, EOV-Seg (L) outperforms ODISE by $+1.1$ PQ with faster speed ($19 \times$), fewer parameters (-1297M), and lower computational complexity (-576GFLOPs).
Compared to FC-CLIP~\cite{yu2023fcclip}, EOV-Seg (L) is about $4$ times faster than it and still shows competitive performance.
It is worth mentioning that compared to other state-of-the-art methods like MasQCLIP~\cite{xu2023masqclip}, OPSNet~\cite{10378471OPSNet}, FreeSeg~\cite{qin2023freeseg}, and MaskCLIP~\cite{ding2023maskclip}, EOV-seg shows significant improvements, such as PQ increased by $+1.0$ to $+9.4$, SQ increased by $+0.6$ to $+17.9$, RQ increased by $+0.9$ to $+10.9$, mIoU increased by $+1.8$ to $+8.4$ and the inference speed is $3.5$ to $19.3$ times faster.
These results demonstrate that our method achieves the optimal trade-off between performance and speed.
\\
\textbf{Task-general EOV-Seg Is All You Need.}
Our method is task-universal, and it can not only perform panoptic segmentation but also semantic segmentation.
%
%
In Tab.\,\ref{tab2:adk20_semantic}, EOV-Seg delivers performance comparable to current state-of-the-art models and it boasts the fastest inference speed among its peers.
Specially, EOV-Seg (L) outperforms CAT-Seg~\cite{cho2024cat-seg} by $+2.0$ mIoU on the A-150 dataset and is $5.3 \times$ faster, has fewer parameters (-265M) and lower FLOPs than CAT-Seg.
%
%
%
These results demonstrate that even though EOV-Seg is trained only on the COCO Panoptic dataset with fewer semantic classes, it successfully adapts to semantic segmentation tasks, generalizes efficiently to unknown categories, achieves accurate segmentation, and runs faster with fewer parameters.
\begin{table}[t]
\setlength\tabcolsep{2.5pt} 
\small
\centering
  \begin{tabular}{c|c|cccccccc}
    \toprule
    Extractor & index & PQ & SQ & RQ & mIoU & FPS$\uparrow$ & Params$\downarrow$ \\
    \midrule
    \multirow{3}{*}{DINOv2} & 0 & 21.9 & 67.3 & 27.3 & 29.5 & 11.7 & 225M\\
     & 5 & 22.4 & 67.7 & 28.8 & 30.1 & 8.9 & 247M\\
     & 11 & 23.1 & 68.2 & 29.4 & 31.2 & 5.8 & 260M\\
    \hline
    \multirow{3}{*}{SAM} & 0 & 22.4 & 68.3 & 27.8 & 30.8 & 11.6 & 225M\\
     & 5 & 23.7 & 70.4 & 28.8 & 31.6 & 8.7 & 247M\\
     & 11 & 24.9 & 71.2 & 29.5 & 32.5 & 5.2 & 260M\\
    \hline
    \multirow{3}{*}{ViT-based CLIP} & 0 & 24.5 & 70.2 & 30.1 & 32.1 & 11.6 & 225M\\
    & 5 & 23.8 & 68.6 & 29.4 & 32.0 & 9.1 & 247M\\
    & 11 & 23.7 & 68.5 & 29.6 & 31.8 & 5.9 & 260M\\
  \bottomrule
\end{tabular}
\caption{Ablation studies on different blocks of different pre-trained models as spatial awareness extractor. index: denotes block index of the pre-trained models.}
\label{tab3:block_num_ablation}
\end{table}
\subsection{Ablation Study}
\textbf{Spatial Awareness Extractor.} In Tab.\,\ref{tab3:block_num_ablation}, we compared the performance and efficiency of using ViT-B versions of different pre-trained models as spatial-aware extractors.
Specifically, when using different blocks of the DINOv2 backbone as spatial awareness extractor, it shows poor performance.
As shown in Fig.\,\ref{fig3} (b), it contains a lot of noise and lacks fine-grained spatial information, we argue the difficulty in learning feature space patterns due to self-supervised pre-training in DINOv2, coupled with the use of unlabeled data, results in diminished instance discrimination and spatial awareness capabilities.
Since SAM is pre-trained on the fine-grained SA-1B~\cite{kirillov2023sam} dataset, it exhibits enhanced fine-grained spatial representation capabilities in its later blocks, progressively improving the segmentation quality.
However, the increase in RQ is modest, likely constrained by its class-agnostic pre-training.
One key finding is that the first ViT-based block demonstrates comparable performance to SAM, with about $2.2\times$ faster inference speed and fewer parameters (-35M).
This is consistent with the observation in Fig.\,\ref{fig3} (d).
This improvement is due to the ability of early self-attention mechanism to process global information, where each patch of the image captures its respective details.
As the number of blocks increases, performance is saturated and has a downward trend. 
To achieve the best trade-off between performance and inference speed, the first block of the ViT-based CLIP is the best choice.
\begin{table}[t]
\centering
\setlength\tabcolsep{4pt} 
\small
\centering
  \begin{tabular}{cc|cccccccc}
    \toprule
    Fusion & VAS & PQ & SQ & RQ & mIoU & FPS$\uparrow$ & Params$\downarrow$ \\
    \midrule
        \multirow{2}{*}{None} & \ding{56} & 22.1 & 67.2 & 27.3 & 30.3 & \textbf{13.8} & \textbf{218M}\\
         & \ding{52} & 22.7 & 67.3 & 28.2 & 31.4 & 13.6 & 219M\\
        \hline
       \multirow{2}{*}{+EAF} & \ding{56} & 23.0 & 68.2 & 28.3 
     & 30.8 & 12.2 & 223M\\
        & \ding{52} & 22.3 & 66.9 & 27.6 
     & 30.4 & 11.8 & 224M\\
    \hline
    \multirow{2}{*}{+SDI} & \ding{56} & 22.5 & 67.1 & 27.8 & 31.1 & 11.4 & 223M\\
    & \ding{52} & 22.9 & 67.4 & 28.5 & 31.5 & 11.0 & 224M\\
    \hline
    \multirow{2}{*}{+TDEE} & \ding{56} & 23.6 & 68.7 & 30.0 & 31.1 & 11.9 & 225M\\
    & \ding{52} & \textbf{24.5} & \textbf{70.2} & \textbf{30.1} & \textbf{32.1} & 11.6 & 225M\\
  \bottomrule
\end{tabular}
\caption{Ablation studies on key components for EOV-Seg. None: a baseline without spatial awareness fusion; EAF: Early Aggregation Fusion; SDI: Separable Dynamic Interaction; TDEE: Two-way Dynamic Embedding Experts; VAS: Vocabulary-Aware Selection;}
\label{tab4:module_ablation}
\end{table}
\begin{table}[t]
\centering
\setlength\tabcolsep{4pt} 
  \begin{tabular}{c|c|cccccc}
    \toprule
    Feat. Inter. & Feat. Scale & PQ & mIoU & FPS$\uparrow$ & Params$\downarrow$ \\
    \midrule
    CA & multi-scale & 24.7 & 32.5 & 3.4 & 304M \\
    CA & single-scale & 23.8 & 30.4 & 6.6 & 282M \\
    DDA & multi-scale & 22.7 & 29.7 & 9.1 & 245M \\
    DDA & single-scale & 24.5 & 32.1 & 11.6 & 225M \\
  \bottomrule
\end{tabular}
\caption{Ablation studies on different feature interaction methods in lightweight decoder with different feature scales. CA: Cross-Attention; DDA: Dynamic Depthwise Attention.}
\label{tab5:feature_interaction}
\end{table}
\begin{table}[t]
\centering
  \small
  \setlength\tabcolsep{2pt} 
  \begin{adjustbox}{max width=0.46\textwidth}
  \begin{tabular}{l|c|cccc}
    \toprule
    Method&Backbone&PQ&FPS$\uparrow$&Params$\downarrow$&FLOPs$\downarrow$\\
    \midrule
    LPSNet~\cite{hong2021lpsnet} & ResNet50 & 7.5 & 10.4 & 74M & 179G\\
    FPSNet (De et al. 2020) & ResNet50 & 10.4 & 12.8 & 124M & 169G\\
    RealTimePan~\cite{hou2020RealTimePan} & ResNet50 & 12.3 & 16.4 & 90M & 160G\\
    YOSO~\cite{hu2023you} & ResNet50 & 14.2 & 23.6 & \textbf{69M} & 138G\\
    EOV-Seg (S) & ResNet50 & \textbf{15.1} & \textbf{23.8} & 71M & \textbf{133G}\\
  \bottomrule
\end{tabular}
  \end{adjustbox}
\caption{Ablation study on lightweight segmentation framework.}
\label{tab6:efficient_seg_comparison}
\end{table}
\begin{figure}[t]
  \centering
  \includegraphics[width=\linewidth]{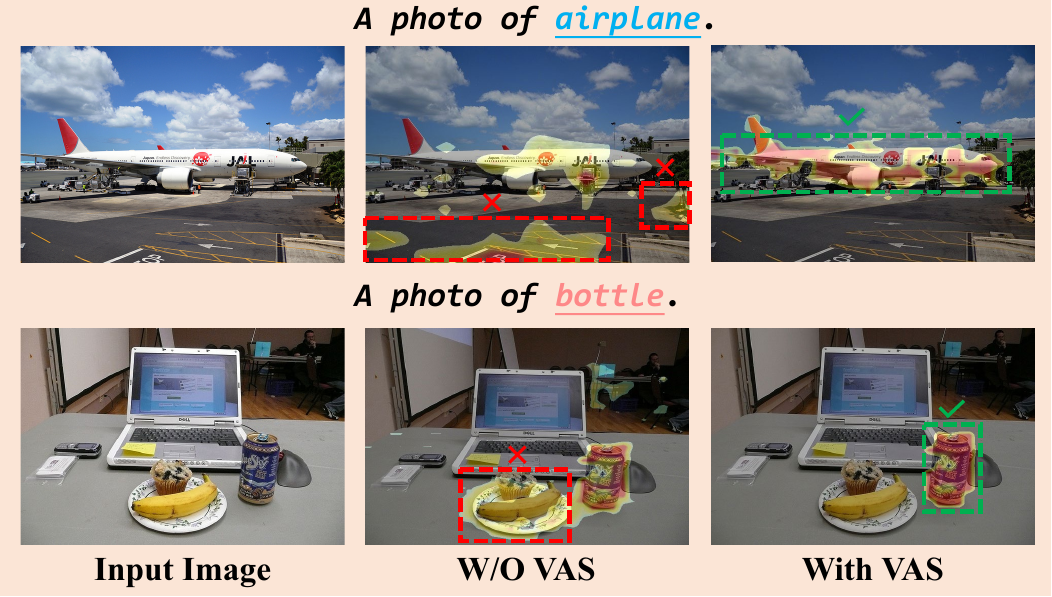}
  \caption{Visualization of similarity map in VAS.}
  \label{fig5}
\end{figure}
\begin{figure}[!htp]
  \centering
  \includegraphics[width=\linewidth]{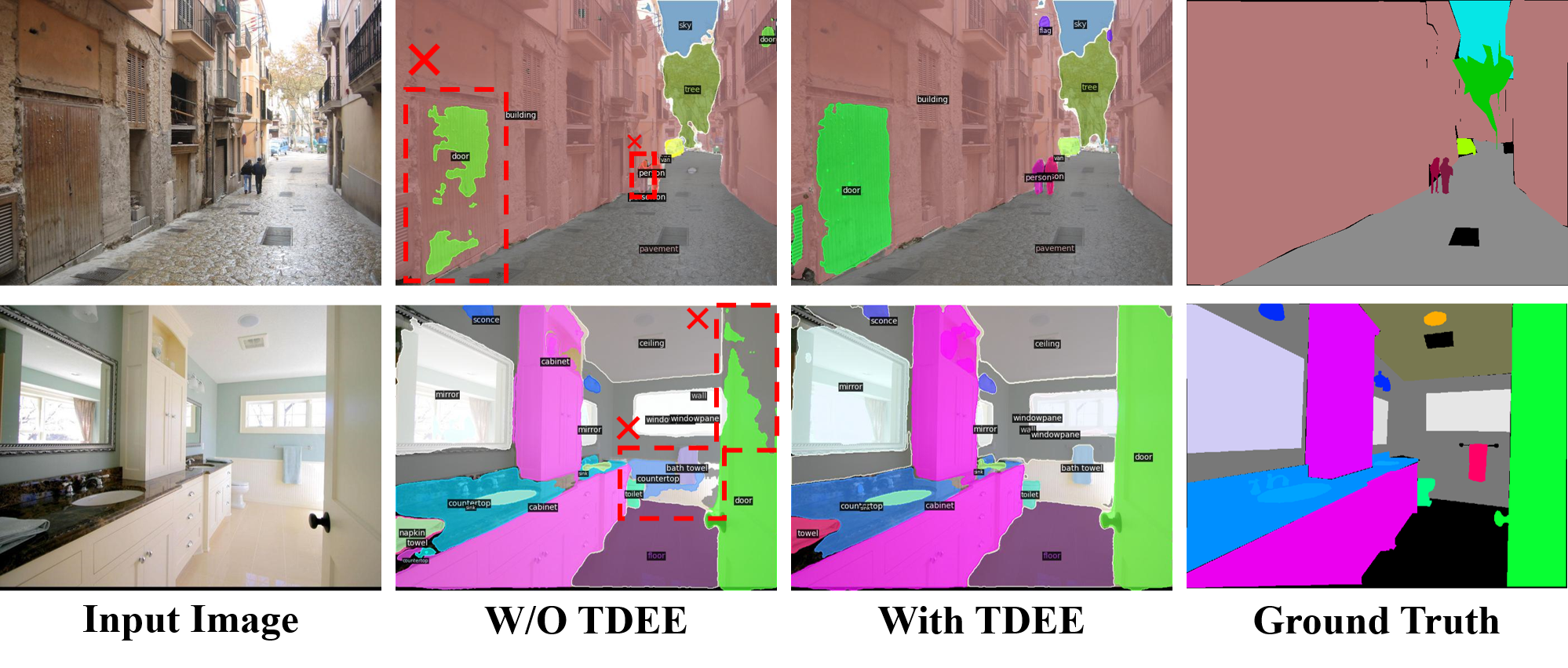}
  \caption{Visualization of segmentation results.}
  \label{fig6}
\end{figure}
\\
\textbf{Our Two Innovative Components.} In Tab.\,\ref{tab4:module_ablation}, we designed two additional variants of spatial awareness fusion, namely EAF and SDI, for more details in the appendix.
All three variants of spatial awareness fusion significantly improved performance, demonstrating the importance of spatial location information for the lightweight decoder.
When we introduce the VAS module for EAF, the performance tends to decline.
This phenomenon is attributed to the early integration of features, which changes the spatial distribution of features and leads to the loss of key spatial information in important semantic features.
However, VAS can cooperate with SDI or TDEE spatial awareness fusion module for mutual benefit and win-win results, and can bring additional performance improvements.
In summary, We finally choose the TDEE and VAS modules, which work together as two innovative components and bring performance improvements of $+2.4$ PQ, $+3.0$ SQ, $+2.8$ RQ and $+1.8$ mIoU without introducing additional computational burden.
\\
\textbf{Feature Interaction Methods.}
In Tab.\,\ref{tab5:feature_interaction}, we explored the performance differences between Cross-Attention~(CA) and Dynamic Depthwise Attention~(DDA) two attention modules and their interaction with features of different scales.
When using CA with multi-scale features only yields marginal improvements over our method, with a $+0.2$ increase in PQ and $+0.4$ in mIoU, but results in a significant slowdown, approximately $3.4\times$ slower in FPS.
Since single scale avoids the fuzzy spatial distribution caused by spatial dimension alignment, and the pre-perception of VAS can reduce the decoder interaction burden, using DDA with single-scale can bring a performance improvement of $+1.8$ PQ and $+2.4$ mIoU over DDA with multi-scale.
\\
\textbf{Efficient Architecture Design.} In Tab.\,\ref{tab6:efficient_seg_comparison}, we compare the performance and efficiency with the previous traditional efficient panoptic segmentation model.
To adapt to the open-vocabulary panoptic segmentation task, we replace the classifier with text embeddings.
EOV-Seg showed superior performance and efficiency, further demonstrating the efficiency and superiority of our architectural design.
\subsection{Qualitative Analysis}
\textbf{Alleviate Mask Decoder Burden.} In Fig.\,\ref{fig5}, we visualize the similarity map between single text embedding and visual-semantic aggregated features.
We found that without VAS, the features may focus on objects with similar colors or nearby positions to the prompt object, confounding feature discrimination.
After VAS enhancement, the features will be more focused on the text-guided area.
\\
\textbf{Infusion of Spatial Awareness.} We visualize the segmentation results with and without Two-way Dynamic Embedding Experts~(TDEE), respectively. 
As shown in the red box in Fig.\,\ref{fig6}, using only the masks generated by the lightweight decoder will result in category confusion and incomplete segmentation.
For instance, the lower body of the "person" merges with the "building" and the segmentation of the "door" is incomplete.
This demonstrates that spatial awareness capabilities is very critical for lightweight decoder.
\textit{See the appendix for more visualizations.}
\section{Conclusion}
\label{sec:conc}
We propose EOV-Seg, an efficient framework designed for open-vocabulary panoptic segmentation.
%
It uses VAS and TDEE to improve instance discrimination capabilities for alleviating the interaction burden on mask decoder and endow mask recognition with spatial awareness capabilities. Extensive experiments show that our EOV-Seg is effective in terms of both accuracy and speed.
\section{Acknowledgments}
This work was supported by National Science and Technology Major Project (No.2022ZD0118201), the National Science Fund for Distinguished Young Scholars (No.62025603), the National Natural Science Foundation of China (No.U21B2037, No.U22B2051, No.U23A20383, No.U21A20472, No.62176222, No.62176223, No.62176226, No.62072386, No.62072387, No.62072389, No.62002305 and No.62272401), and the Natural Science Foundation of Fujian Province of China (No.2021J06003, No.2022J06001).
\bibliography{aaai25}

\begin{thebibliography}{50}
\providecommand{\natexlab}[1]{#1}

\bibitem[{Cavagnero et~al.(2024)Cavagnero, Rosi, Cuttano, Pistilli, Ciccone, Averta, and Cermelli}]{cavagnero2024pem}
Cavagnero, N.; Rosi, G.; Cuttano, C.; Pistilli, F.; Ciccone, M.; Averta, G.; and Cermelli, F. 2024.
\newblock Pem: Prototype-based efficient maskformer for image segmentation.
\newblock In \emph{CVPR}.

\bibitem[{Chen et~al.(2015)Chen, Fang, Lin, Vedantam, Gupta, Doll{\'a}r, and Zitnick}]{chen2015coco_caption}
Chen, X.; Fang, H.; Lin, T.-Y.; Vedantam, R.; Gupta, S.; Doll{\'a}r, P.; and Zitnick, C.~L. 2015.
\newblock Microsoft coco captions: Data collection and evaluation server.
\newblock \emph{arXiv preprint arXiv:1504.00325}.

\bibitem[{Chen et~al.(2023)Chen, Li, Lim, Torralba, and Zhao}]{10378471OPSNet}
Chen, X.; Li, S.; Lim, S.; Torralba, A.; and Zhao, H. 2023.
\newblock Open-vocabulary Panoptic Segmentation with Embedding Modulation.
\newblock In \emph{ICCV}.

\bibitem[{Cheng et~al.(2022)Cheng, Misra, Schwing, Kirillov, and Girdhar}]{cheng2021mask2former}
Cheng, B.; Misra, I.; Schwing, A.~G.; Kirillov, A.; and Girdhar, R. 2022.
\newblock Masked-attention mask transformer for universal image segmentation.
\newblock In \emph{CVPR}.

\bibitem[{Cho et~al.(2024)Cho, Shin, Hong, Arnab, Seo, and Kim}]{cho2024cat-seg}
Cho, S.; Shin, H.; Hong, S.; Arnab, A.; Seo, P.~H.; and Kim, S. 2024.
\newblock Cat-seg: Cost aggregation for open-vocabulary semantic segmentation.
\newblock In \emph{CVPR}.

\bibitem[{Cordts et~al.(2016)Cordts, Omran, Ramos, Rehfeld, Enzweiler, Benenson, Franke, Roth, and Schiele}]{Cordts2016Cityscapes}
Cordts, M.; Omran, M.; Ramos, S.; Rehfeld, T.; Enzweiler, M.; Benenson, R.; Franke, U.; Roth, S.; and Schiele, B. 2016.
\newblock The Cityscapes Dataset for Semantic Urban Scene Understanding.
\newblock In \emph{CVPR}.

\bibitem[{Dai et~al.(2017)Dai, Qi, Xiong, Li, Zhang, Hu, and Wei}]{dai2017deformable_conv}
Dai, J.; Qi, H.; Xiong, Y.; Li, Y.; Zhang, G.; Hu, H.; and Wei, Y. 2017.
\newblock Deformable convolutional networks.
\newblock In \emph{ICCV}.

\bibitem[{Ding et~al.(2022)Ding, Xue, Xia, and Dai}]{ding2021ZegFormer}
Ding, J.; Xue, N.; Xia, G.-S.; and Dai, D. 2022.
\newblock Decoupling zero-shot semantic segmentation.
\newblock In \emph{CVPR}.

\bibitem[{Dosovitskiy et~al.(2021)Dosovitskiy, Beyer, Kolesnikov, Weissenborn, Zhai, Unterthiner, Dehghani, Minderer, Heigold, Gelly, Uszkoreit, and Houlsby}]{dosovitskiy2020vit}
Dosovitskiy, A.; Beyer, L.; Kolesnikov, A.; Weissenborn, D.; Zhai, X.; Unterthiner, T.; Dehghani, M.; Minderer, M.; Heigold, G.; Gelly, S.; Uszkoreit, J.; and Houlsby, N. 2021.
\newblock An Image is Worth 16x16 Words: Transformers for Image Recognition at Scale.
\newblock In \emph{ICLR}.

\bibitem[{Everingham et~al.(2010)Everingham, Van~Gool, Williams, Winn, and Zisserman}]{Everingham10pascalvoc1}
Everingham, M.; Van~Gool, L.; Williams, C. K.~I.; Winn, J.; and Zisserman, A. 2010.
\newblock The Pascal Visual Object Classes (VOC) Challenge.
\newblock \emph{IJCV}.

\bibitem[{Ghiasi et~al.(2022)Ghiasi, Gu, Cui, and Lin}]{ghiasi2021OpenSeg_LSeg+}
Ghiasi, G.; Gu, X.; Cui, Y.; and Lin, T.-Y. 2022.
\newblock Scaling open-vocabulary image segmentation with image-level labels.
\newblock In \emph{ECCV}.

\bibitem[{Gildenblat and contributors(2021)}]{jacobgilpytorch-cam}
Gildenblat, J.; and contributors. 2021.
\newblock PyTorch library for CAM methods.
\newblock \url{https://github.com/jacobgil/pytorch-grad-cam}.

\bibitem[{Gong et~al.(2024)Gong, Zhong, Qu, Luo, Ji, and Jiang}]{gong2024cross}
Gong, Y.; Zhong, Z.; Qu, Y.; Luo, Z.; Ji, R.; and Jiang, M. 2024.
\newblock Cross-modality perturbation synergy attack for person re-identification.
\newblock \emph{arXiv preprint arXiv:2401.10090}.

\bibitem[{Gu et~al.(2021)Gu, Lin, Kuo, and Cui}]{gu2021vild}
Gu, X.; Lin, T.-Y.; Kuo, W.; and Cui, Y. 2021.
\newblock Open-Vocabulary Detection via Vision and Language Knowledge Distillation.
\newblock \emph{arXiv preprint arXiv:2104.13921}.

\bibitem[{Han et~al.(2023)Han, Zhong, Li, Han, and Ma}]{Han2023DeOP}
Han, C.; Zhong, Y.; Li, D.; Han, K.; and Ma, L. 2023.
\newblock Open-vocabulary semantic segmentation with decoupled one-pass network.
\newblock In \emph{ICCV}.

\bibitem[{Hong et~al.(2021)Hong, Guo, Zhang, Chen, and Chu}]{hong2021lpsnet}
Hong, W.; Guo, Q.; Zhang, W.; Chen, J.; and Chu, W. 2021.
\newblock Lpsnet: A lightweight solution for fast panoptic segmentation.
\newblock In \emph{CVPR}.

\bibitem[{Hou et~al.(2020)Hou, Li, Bhargava, Raventos, Guizilini, Fang, Lynch, and Gaidon}]{hou2020RealTimePan}
Hou, R.; Li, J.; Bhargava, A.; Raventos, A.; Guizilini, V.; Fang, C.; Lynch, J.; and Gaidon, A. 2020.
\newblock Real-time panoptic segmentation from dense detections.
\newblock In \emph{CVPR}.

\bibitem[{Hu et~al.(2023)Hu, Huang, Ren, Zhang, Ji, and Cao}]{hu2023you}
Hu, J.; Huang, L.; Ren, T.; Zhang, S.; Ji, R.; and Cao, L. 2023.
\newblock You Only Segment Once: Towards Real-Time Panoptic Segmentation.
\newblock In \emph{CVPR}.

\bibitem[{Jacobs et~al.(1991)Jacobs, Jordan, Nowlan, and Hinton}]{6797059moe}
Jacobs, R.~A.; Jordan, M.~I.; Nowlan, S.~J.; and Hinton, G.~E. 1991.
\newblock Adaptive mixtures of local experts.
\newblock \emph{Neural computation}.

\bibitem[{Jia et~al.(2021)Jia, Yang, Xia, Chen, Parekh, Pham, Le, Sung, Li, and Duerig}]{align}
Jia, C.; Yang, Y.; Xia, Y.; Chen, Y.-T.; Parekh, Z.; Pham, H.; Le, Q.; Sung, Y.-H.; Li, Z.; and Duerig, T. 2021.
\newblock Scaling up visual and vision-language representation learning with noisy text supervision.
\newblock In \emph{ICML}.

\bibitem[{Kirillov et~al.(2023)Kirillov, Mintun, Ravi, Mao, Rolland, Gustafson, Xiao, Whitehead, Berg, Lo, Dollar, and Girshick}]{kirillov2023sam}
Kirillov, A.; Mintun, E.; Ravi, N.; Mao, H.; Rolland, C.; Gustafson, L.; Xiao, T.; Whitehead, S.; Berg, A.~C.; Lo, W.-Y.; Dollar, P.; and Girshick, R. 2023.
\newblock Segment Anything.
\newblock In \emph{ICCV}.

\bibitem[{Li et~al.(2024)Li, Yuan, Li, Ding, Wu, Zhang, Li, Chen, and Loy}]{OMGSeg}
Li, X.; Yuan, H.; Li, W.; Ding, H.; Wu, S.; Zhang, W.; Li, Y.; Chen, K.; and Loy, C.~C. 2024.
\newblock OMG-Seg: Is one model good enough for all segmentation?
\newblock In \emph{CVPR}.

\bibitem[{Liang et~al.(2023)Liang, Wu, Dai, Li, Zhao, Zhang, Zhang, Vajda, and Marculescu}]{liang2023ovseg}
Liang, F.; Wu, B.; Dai, X.; Li, K.; Zhao, Y.; Zhang, H.; Zhang, P.; Vajda, P.; and Marculescu, D. 2023.
\newblock Open-vocabulary semantic segmentation with mask-adapted clip.
\newblock In \emph{CVPR}.

\bibitem[{Lin et~al.(2024)Lin, Shen, Wang, Lin, Li, and Cao}]{WSOVOD_2024_AAAI}
Lin, J.; Shen, Y.; Wang, B.; Lin, S.; Li, K.; and Cao, L. 2024.
\newblock Weakly supervised open-vocabulary object detection.
\newblock In \emph{AAAI}.

\bibitem[{Lin et~al.(2014)Lin, Maire, Belongie, Hays, Perona, Ramanan, Doll{\'a}r, and Zitnick}]{lin2014coco}
Lin, T.-Y.; Maire, M.; Belongie, S.; Hays, J.; Perona, P.; Ramanan, D.; Doll{\'a}r, P.; and Zitnick, C.~L. 2014.
\newblock Microsoft coco: Common objects in context.
\newblock In \emph{ECCV}.

\bibitem[{Mi et~al.(2022)Mi, Lin, Zhou, Shen, Luo, Sun, Cao, Fu, Xu, and Ji}]{mi2022active}
Mi, P.; Lin, J.; Zhou, Y.; Shen, Y.; Luo, G.; Sun, X.; Cao, L.; Fu, R.; Xu, Q.; and Ji, R. 2022.
\newblock Active teacher for semi-supervised object detection.
\newblock In \emph{CVPR}.

\bibitem[{Mottaghi et~al.(2014)Mottaghi, Chen, Liu, Cho, Lee, Fidler, Urtasun, and Yuille}]{6909514pascalvoc2}
Mottaghi, R.; Chen, X.; Liu, X.; Cho, N.-G.; Lee, S.-W.; Fidler, S.; Urtasun, R.; and Yuille, A. 2014.
\newblock The Role of Context for Object Detection and Semantic Segmentation in the Wild.
\newblock In \emph{CVPR}.

\bibitem[{Oquab et~al.(2023)Oquab, Darcet, Moutakanni, Vo, Szafraniec, Khalidov, Fernandez, Haziza, Massa, El-Nouby et~al.}]{oquab2023dinov2}
Oquab, M.; Darcet, T.; Moutakanni, T.; Vo, H.; Szafraniec, M.; Khalidov, V.; Fernandez, P.; Haziza, D.; Massa, F.; El-Nouby, A.; et~al. 2023.
\newblock Dinov2: Learning robust visual features without supervision.
\newblock \emph{arXiv preprint arXiv:2304.07193}.

\bibitem[{Qin et~al.(2023)Qin, Wu, Yan, Li, Yuxi, Xiao, Wang, Wang, Wen, Pan et~al.}]{qin2023freeseg}
Qin, J.; Wu, J.; Yan, P.; Li, M.; Yuxi, R.; Xiao, X.; Wang, Y.; Wang, R.; Wen, S.; Pan, X.; et~al. 2023.
\newblock FreeSeg: Unified, Universal and Open-Vocabulary Image Segmentation.
\newblock In \emph{CVPR}.

\bibitem[{Qu et~al.(2024)Qu, Dai, Li, Lin, Cao, Zhang, and Ji}]{qu2024goi}
Qu, Y.; Dai, S.; Li, X.; Lin, J.; Cao, L.; Zhang, S.; and Ji, R. 2024.
\newblock Goi: Find 3d gaussians of interest with an optimizable open-vocabulary semantic-space hyperplane.
\newblock In \emph{ACM MM}.

\bibitem[{Qu, Wang, and Qi(2023)}]{qu2023sg}
Qu, Y.; Wang, Y.; and Qi, Y. 2023.
\newblock Sg-nerf: Semantic-guided point-based neural radiance fields.
\newblock In \emph{ICME}.

\bibitem[{Radford et~al.(2021)Radford, Kim, Hallacy, Ramesh, Goh, Agarwal, Sastry, Askell, Mishkin, Clark et~al.}]{Radford2021CLIP}
Radford, A.; Kim, J.~W.; Hallacy, C.; Ramesh, A.; Goh, G.; Agarwal, S.; Sastry, G.; Askell, A.; Mishkin, P.; Clark, J.; et~al. 2021.
\newblock Learning transferable visual models from natural language supervision.
\newblock In \emph{ICML}.

\bibitem[{Rasheed et~al.(2023)Rasheed, Khattak, Maaz, Khan, and Khan}]{hanoona2023vificlip}
Rasheed, H.; Khattak, M.~U.; Maaz, M.; Khan, S.; and Khan, F.~S. 2023.
\newblock Fine-tuned clip models are efficient video learners.
\newblock In \emph{CVPR}.

\bibitem[{Ren et~al.(2016)Ren, He, Girshick, and Sun}]{renNIPS15fasterrcnn}
Ren, S.; He, K.; Girshick, R.; and Sun, J. 2016.
\newblock Faster R-CNN: Towards real-time object detection with region proposal networks.
\newblock \emph{TPAMI}.

\bibitem[{Rombach et~al.(2022)Rombach, Blattmann, Lorenz, Esser, and Ommer}]{stable_diffusion}
Rombach, R.; Blattmann, A.; Lorenz, D.; Esser, P.; and Ommer, B. 2022.
\newblock High-Resolution Image Synthesis With Latent Diffusion Models.
\newblock In \emph{CVPR}.

\bibitem[{VS et~al.(2024)VS, Borse, Park, Das, Patel, Hayat, and Porikli}]{vs2024possam}
VS, V.; Borse, S.; Park, H.; Das, D.; Patel, V.; Hayat, M.; and Porikli, F. 2024.
\newblock Possam: Panoptic open-vocabulary segment anything.
\newblock \emph{arXiv preprint arXiv:2403.09620}.

\bibitem[{Weng et~al.(2023)Weng, Yang, Li, Wu, and Jiang}]{weng2023Open-VCLIP}
Weng, Z.; Yang, X.; Li, A.; Wu, Z.; and Jiang, Y.-G. 2023.
\newblock Open-vclip: Transforming clip to an open-vocabulary video model via interpolated weight optimization.
\newblock In \emph{ICML}.

\bibitem[{Wu et~al.(2023)Wu, Zhu, Zhao, and Li}]{wu2023cora}
Wu, X.; Zhu, F.; Zhao, R.; and Li, H. 2023.
\newblock Cora: Adapting clip for open-vocabulary detection with region prompting and anchor pre-matching.
\newblock In \emph{CVPR}.

\bibitem[{Xu et~al.(2022{\natexlab{a}})Xu, De~Mello, Liu, Byeon, Breuel, Kautz, and Wang}]{xu2022groupvit}
Xu, J.; De~Mello, S.; Liu, S.; Byeon, W.; Breuel, T.; Kautz, J.; and Wang, X. 2022{\natexlab{a}}.
\newblock Groupvit: Semantic segmentation emerges from text supervision.
\newblock In \emph{CVPR}.

\bibitem[{Xu et~al.(2023{\natexlab{a}})Xu, Liu, Vahdat, Byeon, Wang, and De~Mello}]{xu2023odise}
Xu, J.; Liu, S.; Vahdat, A.; Byeon, W.; Wang, X.; and De~Mello, S. 2023{\natexlab{a}}.
\newblock Open-vocabulary panoptic segmentation with text-to-image diffusion models.
\newblock In \emph{CVPR}.

\bibitem[{Xu et~al.(2023{\natexlab{b}})Xu, Zhang, Wei, Hu, and Bai}]{xu2023SAN}
Xu, M.; Zhang, Z.; Wei, F.; Hu, H.; and Bai, X. 2023{\natexlab{b}}.
\newblock Side adapter network for open-vocabulary semantic segmentation.
\newblock In \emph{CVPR}.

\bibitem[{Xu et~al.(2022{\natexlab{b}})Xu, Zhang, Wei, Lin, Cao, Hu, and Bai}]{xu2021simbaseline}
Xu, M.; Zhang, Z.; Wei, F.; Lin, Y.; Cao, Y.; Hu, H.; and Bai, X. 2022{\natexlab{b}}.
\newblock A simple baseline for open-vocabulary semantic segmentation with pre-trained vision-language model.
\newblock In \emph{ECCV}.

\bibitem[{Xu et~al.(2023{\natexlab{c}})Xu, Xiong, Ding, and Tu}]{xu2023masqclip}
Xu, X.; Xiong, T.; Ding, Z.; and Tu, Z. 2023{\natexlab{c}}.
\newblock MasQCLIP for Open-Vocabulary Universal Image Segmentation.
\newblock In \emph{ICCV}.

\bibitem[{Yu et~al.(2023)Yu, He, Deng, Shen, and Chen}]{yu2023fcclip}
Yu, Q.; He, J.; Deng, X.; Shen, X.; and Chen, L.-C. 2023.
\newblock Convolutions die hard: Open-vocabulary segmentation with single frozen convolutional clip.
\newblock In \emph{NeurIPS}.

\bibitem[{Yue et~al.(2024)Yue, Lin, Zhang, Hu, Lu, Niu, Ding, Zhang, Jiang, Cao et~al.}]{yue2024adaptive}
Yue, P.; Lin, J.; Zhang, S.; Hu, J.; Lu, Y.; Niu, H.; Ding, H.; Zhang, Y.; Jiang, G.; Cao, L.; et~al. 2024.
\newblock Adaptive Selection based Referring Image Segmentation.
\newblock In \emph{ACM MM}.

\bibitem[{Zhang et~al.(2022)Zhang, Guo, Zhang, Li, Miao, Cui, Qiao, Gao, and Li}]{zhang2021pointclip}
Zhang, R.; Guo, Z.; Zhang, W.; Li, K.; Miao, X.; Cui, B.; Qiao, Y.; Gao, P.; and Li, H. 2022.
\newblock Pointclip: Point cloud understanding by clip.
\newblock In \emph{CVPR}.

\bibitem[{Zheng~Ding(2023)}]{ding2023maskclip}
Zheng~Ding, Z.~T., Jieke~Wang. 2023.
\newblock Open-Vocabulary Universal Image Segmentation with MaskCLIP.
\newblock In \emph{ICML}.

\bibitem[{Zhong et~al.(2022)Zhong, Yang, Zhang, Li, Codella, Li, Zhou, Dai, Yuan, Li et~al.}]{zhong2022regionclip}
Zhong, Y.; Yang, J.; Zhang, P.; Li, C.; Codella, N.; Li, L.~H.; Zhou, L.; Dai, X.; Yuan, L.; Li, Y.; et~al. 2022.
\newblock Regionclip: Region-based language-image pretraining.
\newblock In \emph{CVPR}.

\bibitem[{Zhou et~al.(2017)Zhou, Zhao, Puig, Fidler, Barriuso, and Torralba}]{zhou2017ade20k}
Zhou, B.; Zhao, H.; Puig, X.; Fidler, S.; Barriuso, A.; and Torralba, A. 2017.
\newblock Scene parsing through ade20k dataset.
\newblock In \emph{CVPR}.

\bibitem[{Zhu et~al.(2023)Zhu, Zhang, He, Guo, Zeng, Qin, Zhang, and Gao}]{Zhu2022PointCLIPV2}
Zhu, X.; Zhang, R.; He, B.; Guo, Z.; Zeng, Z.; Qin, Z.; Zhang, S.; and Gao, P. 2023.
\newblock Pointclip v2: Prompting clip and gpt for powerful 3d open-world learning.
\newblock In \emph{ICCV}.

\end{thebibliography}

\clearpage
\appendix

\section{Overview}
In the following supplementary materials, we elaborate on the inference details, provide the pseudocode of the Vocabulary-Aware Selection (VAS) module, and delineate the pipeline of the first two fusion methods of spatial awareness fusion, namely Early Aggregation Fusion (EAF) and Separable Dynamic Interaction (SDI).
In addition, we present the additional experimental results and additional qualitative results, respectively.

\section{Inference Details}
\label{supple:inference}
During inference, \(C_{test}\) may include categories not present in \(C_{train}\).
To enhance the model's generalization performance on unseen classes, we leverage the powerful open-vocabulary recognition capability of the pre-trained CLIP~\cite{Radford2021CLIP}.
Specifically, we perform mask pooling on the last layer features of the CNN-based CLIP backbone with the predicted masks to obtain the class embedding \(E_{c}\in\mathbb{R}^{N_{mask}\times D}\).
The out-of-vocabulary classification scores \(S_O\) is obtained by computing the dot product between the class embedding \(E_c\) and text embedding \(E_t\).
Then, we employ geometric ensemble~\cite{gu2021vild, ghiasi2021OpenSeg_LSeg+, xu2023odise, yu2023fcclip} to fuse in-vocabulary and out-of-vocabulary classification scores as obtain final classification scores \(S\in\mathbb{R}^{N_{mask}\times N_{class}}\):
\begin{align}
    S_{i}=
    \begin{cases}
        S_{I, i}^{(1-\alpha)}\cdot S_{O, i}^\alpha~, &\text{if}~i\in C_{train} \\
        S_{I, i}^{(1-\beta)}\cdot S_{O, i}^\beta~, &\text{otherwise}
    \end{cases}
    \label{eq:ensemble}
\end{align}
where \(i\) denotes category index, \(S_{I, i}\) and \(S_{O, i}\) is the \(i^{th}\) in-vocabulary classification score and out-of-vocabulary classification score, respectively.
\(\alpha\) and \(\beta\) are fixed balancing factors of the seen and unseen categories, respectively.
We set \(\alpha=0.4\) and \(\beta=0.8\) across all datasets.
\section{Vocabulary-Aware Selection}
\label{supple:vas}
As illustrated in Algorithm~\ref{alg:method_pseudocode}, we briefly present the pseudocode of the Vocabulary-Aware Selection (VAS) module in PyTorch style.
We use \textit{bs}, \textit{n} and \textit{ndim} for batch size, number~of~text~prompts and feature dimension in the pseudo-code, respectively.
The variables \textit{agg\_feat} and \textit{text\_embed} are used for visual aggregated features \(F_{agg}\) and text embeddings \(E_t\), respectively.
They are mapped to the same feature dimension through linear layers.
Then we divide them into multi-head visual aggregated features (\textit{mh\_agg\_feat}) and multi-head text embedding (\textit{mh\_text\_embed}) along the channel dimension.
We use \textit{m} and \textit{c} for number~of~heads and number~of~channels~per~head.
Subsequently, we compute vocabulary-aware~attention~weights (\textit{attn\_weights}) following pseudo-code and we obtain visual-semantic~aggregated~features (vs\_agg\_feat) by element-wise multiplication of \textit{agg\_feat} and \textit{attn\_weights}.
\definecolor{mygreen}{RGB}{0, 128, 0}
\begin{algorithm}[t]
    \caption{Vocabulary-Aware Selection}
    \label{alg:method_pseudocode}
    \begin{algorithmic}
        \State \textcolor{mygreen}{\texttt{"""}} 
        \State \textcolor{mygreen}{\textbf{Input:}}
        \State \textcolor{mygreen}{\hspace{\algorithmicindent} $agg\_feat$: $(bs,~ndim,~h/4,~w/4)$}
        \State \textcolor{mygreen}{\hspace{\algorithmicindent} $text\_embed$: $(bs,~n,~ndim)$}
        \State \textcolor{mygreen}{\textbf{Output:}}
        \State \textcolor{mygreen}{\hspace{\algorithmicindent} $vs\_agg\_feat$: $(bs,~ndim,~h/4,~w/4)$}
        \State \textcolor{mygreen}{\texttt{"""}}
        \State
        \State agg\_feat\_proj $=$ DepthwiseSeparableConv2d(agg\_feat)
        \State text\_embed\_proj $=$ Linear(text\_embed) \\
        \State \textcolor{mygreen}{$\#~Split~Into~Multi Heads.$}
        \State mh\_agg\_feat $=$ agg\_feat\_proj.reshape(bs, m, c, h, w)
        \State mh\_text\_embed $=$ text\_embed\_proj.reshape(bs, -1, m, c) \\
        \State \textcolor{mygreen}{$\#~Compute~Vocabulary-Aware~Attention~Weights.$}
        \State attn\_weights=torch.einsum('bmchw,bnmc$\rightarrow$bmhwn', mh\_agg\_feat, mh\_text\_embed)
        \State attn\_weights $=$ Softmax(attn\_weights, dim=-1)
        \State attn\_weights $=$ attn\_weights.max(-1)[0]
        \State attn\_weights $=$ attn\_weights $\ast$ scale + offset
        \State agg\_feat $=$ agg\_feat.reshape(bs, m, -1, h, w) \\
        \State \textcolor{mygreen}{$\#~Output~Visual-Semantic~Aggregated~Features.$}
        \State vs\_agg\_feat $=$ agg\_feat $\ast$ attn\_weights
    \end{algorithmic}
\end{algorithm}
\section{Spatial Awareness Fusion}
\label{supple:saf}
\begin{figure}[t]
  \centering
  \includegraphics[width=0.8\linewidth]{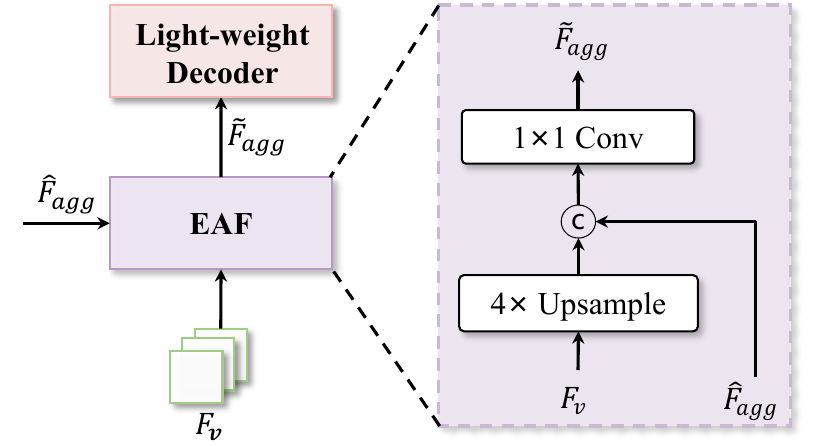}
  \caption{Overview of Early Aggregation Fusion.}
  \label{supple:variant1}
\end{figure}

In this section, we provide detailed overview diagrams of two spatial fusion modules, \textit{i.e.}, Early Aggregation Fusion (EAF) and Separable Dynamic Interaction (SDI).
\\
\textbf{Early Aggregation Fusion.}
As illustrated in Fig.~\ref{supple:variant1}, the visual-semantic aggregated features \(\hat{F}_{agg}\) and the visual-spatial features \(F_v\) with reshape and 4\(\times\) upsampling are concatenated along the channel dimension.
Then, they are fused through a \(1\times 1\) convolution to obtain the aggregated features with instance and spatial awareness capabilities \(\hat{F}_{agg}\in\mathbb{R}^{D\times \frac{H}{4}\times \frac{W}{4}}\).
This design injects auxiliary instance awareness and spatial location information into the aggregated features during the early stages of decoding. 
This allows the aggregated features to roughly perceive the position and scale of objects at the decoder layer, which is beneficial for the object kernels to precisely locate and segment everything.
\begin{figure}[t]
  \centering
  \includegraphics[width=0.8\linewidth]{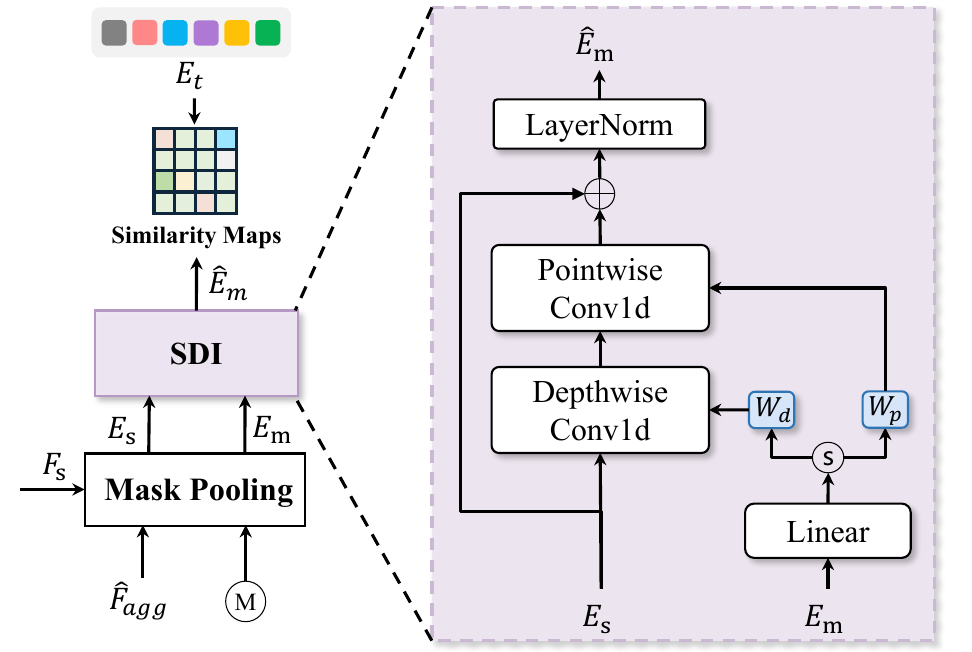}
  \caption{Overview of Separable Dynamic Interaction.}
  \label{supple:variant2}
\end{figure}
\\
\textbf{Separable Dynamic Interaction.}
As illustrated in Fig.~\ref{supple:variant2}, we take the mask embedding \(E_m\) as the query to dynamically generate weights for depthwise convolution and pointwise convolution.
Then we take the spatial awareness embedding \(E_s\) as the value, and use depthwise and pointwise convolutions to extract spatial location information on the spatial awareness embedding \(E_s\).
This design benefits from the convolutional interaction between the mask embedding \(E_m\) and the spatial awareness embedding \(E_s\), and can inject instance semantic and spatial location information into the mask embedding \(E_m\) while maintaining a lower parameter count.
\section{Additional Experimental Results}
\label{supple:exp}
\textbf{Closed-Vocabulary and Open-Vocabulary panoptic segmentation performance.} In Tab.~\ref{supple:coco_tab}, we compared EOV-Seg with SOTAs on COCO panoptic segmentation dataset to evaluate the closed-vocabulary capabilities.
Specifically, with only a small increase in the number of parameters, the performance of EOV-Seg significantly surpassed FreeSeg~\cite{qin2023freeseg}, with PQ improved by $+15.5$, SQ improved by $+2.2$, RQ improved by $+18.4$, and mIoU improved by +18 and FPS is about $15\times$ faster.
Additionally, EOV-Seg outperforms ODISE~\cite{xu2023odise} by $+1.3$ PQ, $+7.8$ mIoU, and the inference speed is $17\times$ faster with fewer parameters (-1297M).
These results demonstrate that EOV-Seg not only has good generalization performance in open-vocabulary scenes, but also achieves excellent performance in closed-vocabulary scenes.
%
In Tab.~\ref{supple:cityscapes_tab}, we present open-vocabulary panoptic segmentation results on the Cityscapes~\cite{Cordts2016Cityscapes} dataset. 
EOV-Seg surpasses ODISE~\cite{xu2023odise}, PQ is improved by $+10.8$, RQ is improved by $+13.9$ with fewer parameters (-1297M).
Even under challenging street scene datasets, EOV-Seg has strong open-vocabulary recognition and segmentation capabilities.
\\
\textbf{Input image at different scales.} In Tab.~\ref{supple:scales_tab}, the PQ and mIoU of the input image scale is (640, 2560) are higher than (800, 1333), and the FPS is faster.
\\
\textbf{Different ensemble methods and hyperparameters.} In Tab.~\ref{supple:hyperparam_tab}, we evaluate the effectiveness of different hyperparameter combinations under different ensemble methods on the ADE20K dataset. 
Specifically, we evaluate arithmetic ensemble and geometric ensemble, respectively.
Arithmetic ensemble is a linear weighted ensemble. 
Geometric ensemble is defined as shown in Eq.~(\ref{eq:ensemble}). 
Both show a consistent trend in EOV-Seg, but the geometric ensemble has more advantages than the arithmetic ensemble, and when the hyperparameter \(\alpha\) of the geometric ensemble is equal to 0.4 and \(\beta\) is equal to 0.8, EOV-Seg has the best performance. 
This indicates that for seen classes, the preference is to use in-vocabulary classification, and for unseen classes, the preference is to use out-of-vocabulary classification.
\begin{table}[!htp]
\centering
    \begin{adjustbox}{max width=0.47\textwidth}
    \begin{tabular}{c|ccccccc}
    \toprule
    Method & PQ & SQ & RQ & mIoU & FPS & Params \\
    \midrule
    FreeSeg~\cite{qin2023freeseg} & 31.4 & 78.3 & 38.9 & 42.2 & 0.8 & \textbf{219M} \\
    ODISE~\cite{xu2023odise} & 45.6 & - & - & 52.4 & 0.7 & 1522M \\
    EOV-Seg (L) & \textbf{46.9} & \textbf{80.5} & \textbf{57.3} & \textbf{60.2} & \textbf{11.9} & 225M \\
    \bottomrule
    \end{tabular}
    \end{adjustbox}
    \caption{Closed-vocabulary panoptic segmentation performance on COCO~\cite{lin2014coco} dataset.}
    \label{supple:coco_tab}
\end{table}
\begin{table}[!htp]
\centering
\begin{adjustbox}{max width=0.47\textwidth}
    \begin{tabular}{c|ccccccc}
    \toprule
    Method & PQ & SQ & RQ & mIoU & FPS & Params \\
    \midrule
    ODISE~\cite{xu2023odise} & 23.9 & \textbf{75.3} & 29.0 & - & - & 1522M \\
    EOV-Seg (L) & \textbf{34.7} & 73.9 & \textbf{42.9} & 50.2 & 7.9 & \textbf{225M} \\
    \bottomrule
    \end{tabular}
\end{adjustbox}
    \caption{Open-vocabulary panoptic segmentation performance on Cityscapes~\cite{Cordts2016Cityscapes} dataset. ODISE~\cite{xu2023odise} does not provide a test script on Cityscapes~\cite{Cordts2016Cityscapes} dataset.}
    \label{supple:cityscapes_tab}
\end{table}
\begin{table}[!htp]
\centering
    \begin{adjustbox}{max width=0.47\textwidth}
    \begin{tabular}{c|cccccc}
    \toprule
    Method & Scales & PQ & SQ & RQ & mIoU & FPS\\
    \midrule
    \multirow{2}{*}{EOV-Seg (S)} & (800, 1333) & 14.3 & 54.1 & 17.7 & 20.4 & 16.4 \\
    & (640, 2560) & 15.1 & 57.0 & 18.9 & 21.9 & 24.5 \\
    \hline
    \multirow{2}{*}{EOV-Seg (L)} & (800, 1333) & 23.1 & 70.2 & 28.3 & 31.3 & 9.1 \\
    & (640, 2560) & 24.5 & 70.2 & 30.1 & 32.1 & 11.6 \\
    \bottomrule
    \end{tabular}
    \end{adjustbox}
    \caption{Comparison of input image at different scales on ADE20K~\cite{zhou2017ade20k} dataset. (640, 2560) denotes the shortest side of input images is resized to 640 pixels, while ensuring the longer side does not exceed 2560 pixels.}
    \label{supple:scales_tab}
\end{table}
\begin{table}[!htp]
\centering
  \begin{tabular}{c|c|c|c}
    \toprule
    \(\alpha\) & \(\beta\) & Arithmetic & Geometric \\
    \midrule
    0.0 & 0.0 & 15.4 & 15.4 \\
    0.2 & 0.8 & 22.4 & 22.7 \\
    0.4 & 0.6 & 22.8 & 23.1 \\
    \textbf{0.4} & \textbf{0.8} & \textbf{23.4} & \textbf{24.5} \\
    0.5 & 0.5 & 22.6 & 22.8 \\
    0.8 & 0.4 & 21.8 & 21.9 \\
    1.0 & 1.0 & 20.2 & 20.2 \\
  \bottomrule
\end{tabular}
\caption{Effectiveness (PQ) of hyperparameters \(\alpha\) and \(\beta\) under different ensemble methods on ADE20K~\cite{zhou2017ade20k} dataset. Arithmetic ensemble is a linear weighted ensemble. Geometric ensemble is defined as shown in Eq.~(\ref{eq:ensemble}).}
\label{supple:hyperparam_tab}
\end{table}
\begin{figure*}[t]
  \centering
  \includegraphics[width=\textwidth]{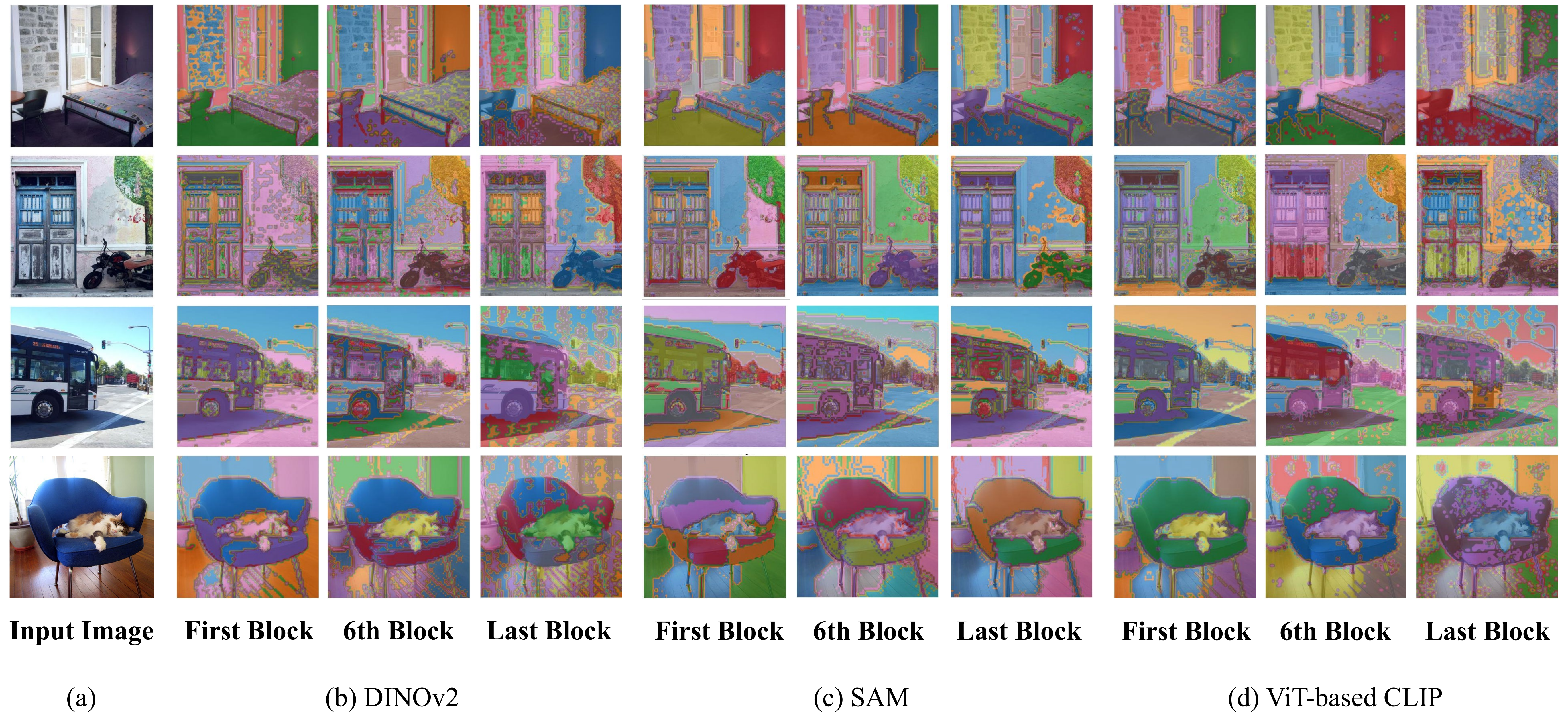}
  \caption{Visualization of K-means clustering of backbone features concerning different blocks across various VFMs (e.g. DINOv2~\cite{oquab2023dinov2}, SAM~\cite{kirillov2023sam}, ViT-based CLIP~\cite{Radford2021CLIP}).}
  \label{supple:k-means_fig}
\end{figure*}
\section{Additional Qualitative Results}
\label{supple:visual}
\begin{figure*}[t]
  \centering
  \includegraphics[width=\linewidth]{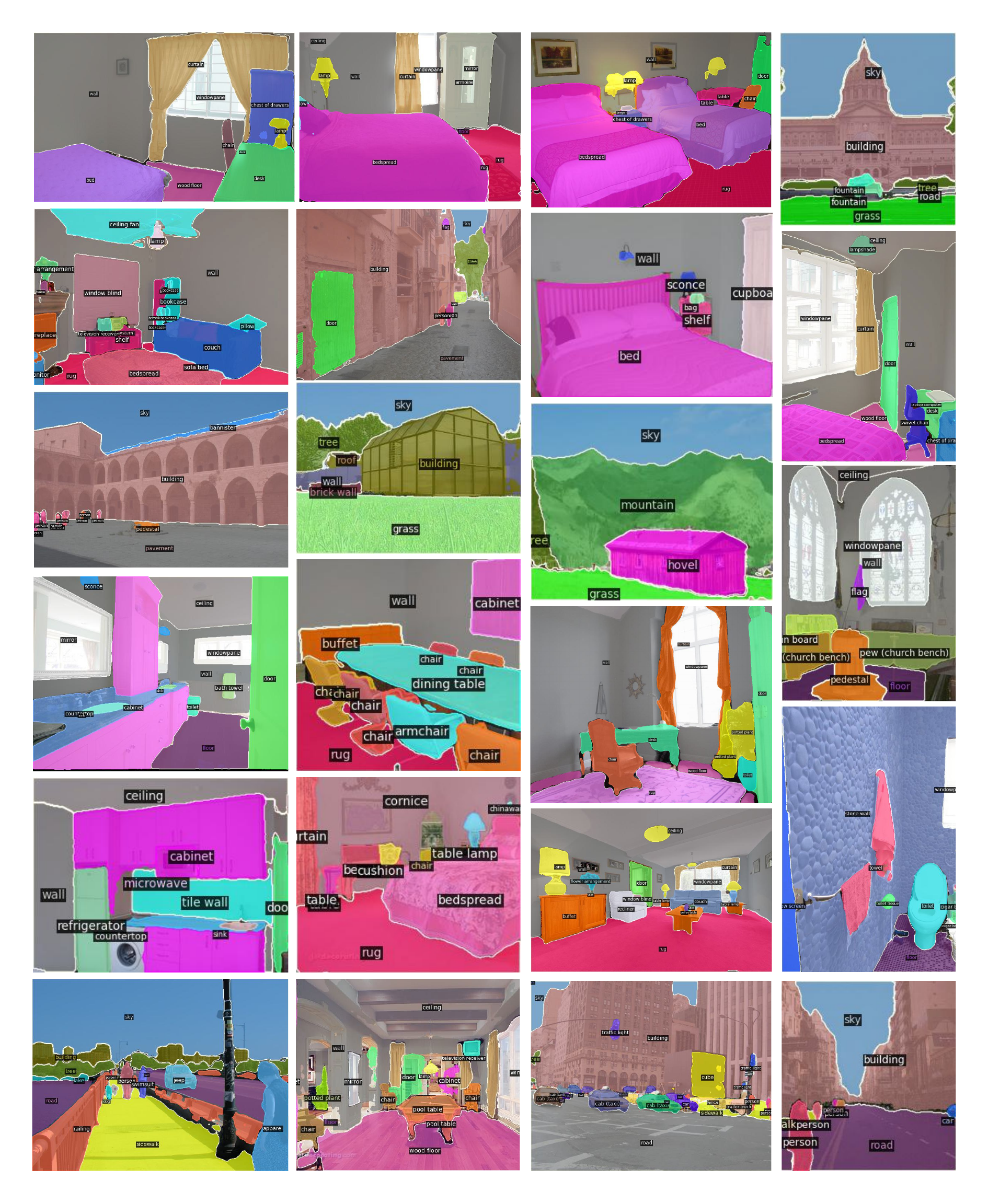}
  \caption{Qualitative visualization of open-vocabulary panoptic segmentation results on ADE20K.}
  \label{supple:ade20k_fig}
\end{figure*}

\begin{figure*}[t]
  \centering
  \includegraphics[width=\linewidth]{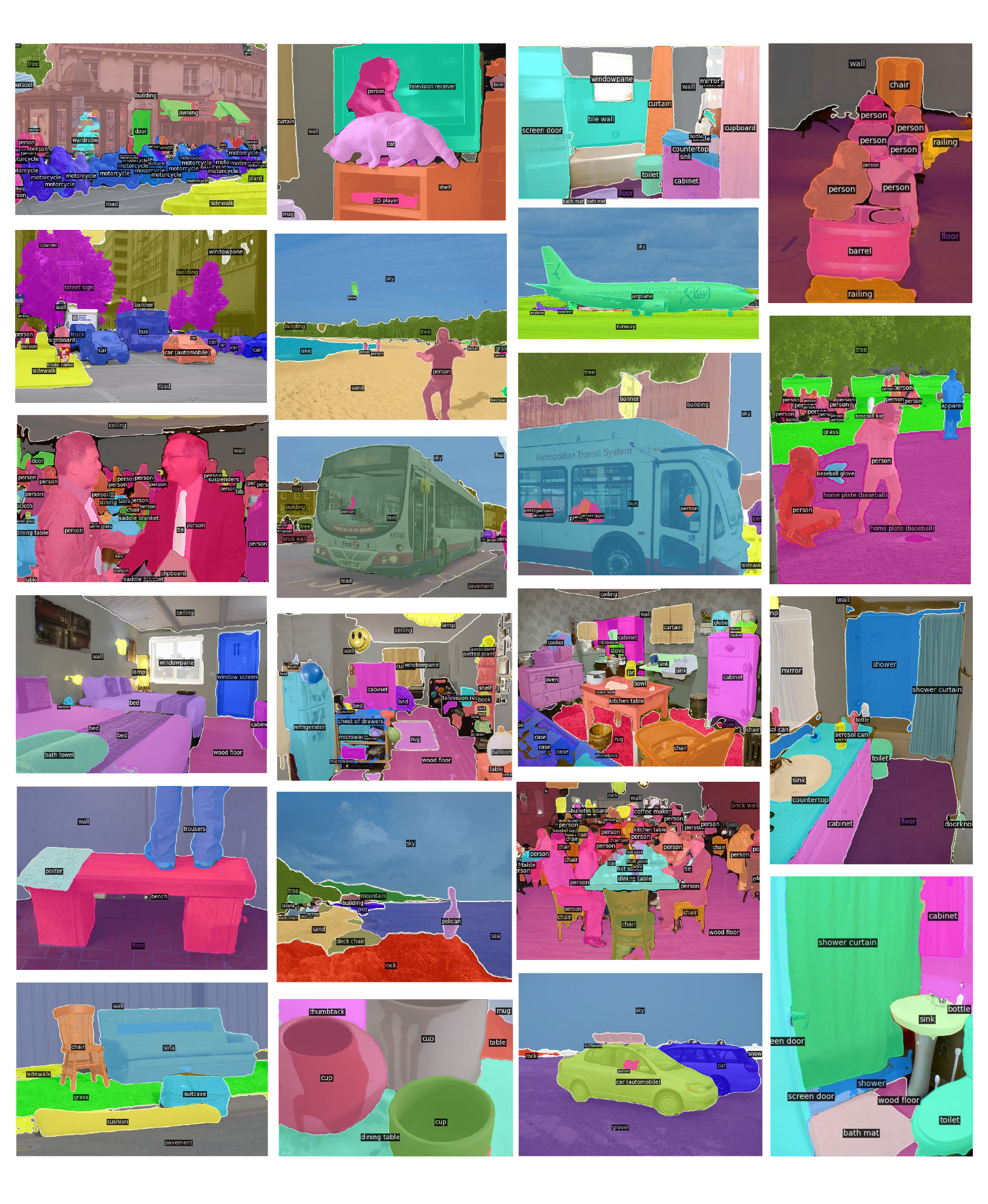}
  \caption{Qualitative visualization of open-vocabulary panoptic segmentation results on COCO.}
  \label{supple:coco_fig}
\end{figure*}

%
In Fig.~\ref{supple:k-means_fig}, we present the visualization of feature clustering of the first block, sixth block, and last block of different VFMs, respectively.
This is consistent with our conclusion in Sec. 1 of the main paper.
DINOv2 is biased towards instance discrimination but has weaker capabilities in spatial awareness.
The last block of SAM has the best fine-grained positioning capabilities, the first block of ViT-based CLIP shows spatial awareness capabilities comparable to SAM, but as the number of blocks increases, feature clustering becomes noisier.
To demonstrate open-vocabulary panoptic segmentation capabilities of EOV-Seg, we perform extensive visualizations of open-vocabulary panoptic segmentation predictions on the ADE20K and COCO dataset in Fig.~\ref{supple:ade20k_fig} and Fig.~\ref{supple:coco_fig}, respectively.

\end{document}